\documentclass{egpubl}

\ConferenceSubmission   
\usepackage[T1]{fontenc}
\usepackage{dfadobe}  

\usepackage{cite}  
\BibtexOrBiblatex
\electronicVersion
\PrintedOrElectronic
\ifpdf \usepackage[pdftex]{graphicx} \pdfcompresslevel=9
\else \usepackage[dvips]{graphicx} \fi

\usepackage{egweblnk} 

\usepackage{pgf}
\usepackage{array}
\usepackage{booktabs} 
\usepackage[noend]{algpseudocode}
\usepackage{algorithm}
\usepackage{varwidth}
\usepackage{amsmath}
\usepackage{amssymb}
\usepackage{MnSymbol}
\usepackage{multicol}
\usepackage{float}
\usepackage{subfigure}
\usepackage{siunitx}
\usepackage{tikz}
\usepackage{nicefrac}
\usepackage{tabularx}
\newcolumntype{L}[1]{>{\raggedright\arraybackslash}p{#1}}
\newcolumntype{C}[1]{>{\centering\arraybackslash}p{#1}}
\newcolumntype{R}[1]{>{\raggedleft\arraybackslash}p{#1}}
\usepackage{stmaryrd}

\title[NeuralQAAD]{NeuralQAAD: An Efficient Differentiable Framework for High Resolution Point Cloud Compression}

\author[Wagner, Schwanecke]
       {\textbf{Nicolas Wagner, Ulrich Schwanecke}\\
         University of Applied Science RheinMain, Germany\\
       }
\begin{document}
\maketitle

\begin{abstract}
In this paper, we propose NeuralQAAD, a differentiable point cloud compression framework that is fast, robust to sampling, and applicable to high resolutions. Previous work that is able to handle complex and non-smooth topologies is hardly scaleable to more than just a few thousand points. We tackle the task with a novel neural network architecture characterized by weight sharing and autodecoding. Our architecture uses parameters much more efficiently than previous work, allowing us to be deeper and scalable. Futhermore, we show that the currently only tractable training criterion for point cloud compression, the Chamfer distance, performances poorly for high resolutions. To overcome this issue, we pair our architecture with a new training procedure based upon a quadratic assignment problem (QAP) for which we state two approximation algorithms. We solve the QAP in parallel to gradient descent. This procedure acts as a surrogate loss and allows to implicitly minimize the more expressive Earth Movers Distance (EMD) even for point clouds with way more than $10^6$ points. As evaluating the EMD on high resolution point clouds is intractable, we propose a divide-and-conquer approach based on k-d trees, the EM-kD, as a scaleable and fast but still reliable upper bound for the EMD. NeuralQAAD is demonstrated on COMA, D-FAUST, and Skulls to significantly outperform the current state-of-the-art visually and in terms of the EM-kD. Skulls is a novel dataset of skull CT-scans which we will make publicly available together with our implementation of NeuralQAAD.

\end{abstract}

\section{Introduction}\label{sec_int}
In recent years, deep learning has been successfully applied to numerous computer vision and graphics tasks, such as classification, segmentation, or compression. However, most progress has been achieved within the 2D domain. Unfortunately, the same methods cannot easily be generalized to unstructured three-di\-men\-sio\-nal data. Memory consumption and computational cost usually increase rapidly in 3D, which limits the applicability of GPU-accelerated algorithms. In particular, gradient descent algorithms suffer from this phenomenon. Furthermore, most learning-based 2D algorithms work on regular data structures like images. Irregular representations such as point clouds or meshes are only sparely covered so far. However, as the typical output of 3D scanners, high resolution point clouds are widely used in autonomous driving, robotics, and other domains to represent surfaces or volumes. A tool capable of reducing the memory requirements of point clouds that can be added to a differential pipeline is therefore highly desirable. Unfortunately, point clouds pose a particular challenge since the points may not only be irregularly but also sparsely scattered in space and, in contrast to meshes, do not provide explicit neighborhood information.
Furthermore, the habit of neural networks to generate smooth signals seems to hinder the reconstruction of high frequencies, complex topologies, and discontinuities.

With autoencoders, deep learning offers a widely used nonlinear tool for compressing data. The straightforward adaption of standard 2D autoencoders to point clouds requires voxelization. Voxelization, however, usually results in a loss of precision and does not exploit sparseness. The first deep learning architecture for directly processing point clouds was PointNet \cite{point}. It is based on extracting global features by max-pooling pointwise local features. Nonetheless, the construction of a point cloud autoencoder requires different or additional concepts since there is no trivial pseudo-invertible counterpart for max-pooling. 

The current state-of-the-art in differentiable point cloud compression mainly focuses on the idea of folding one or multiple fixed input manifolds into a target manifold \cite{ChenDYLFT20, groueix2018, DeprelleGFKRA19}.  Regardless of the respective concept, the size of the point clouds that can be processed by current hardware is very limited, since mainly pointwise neural networks are used. Recently, sampling techniques have been used to overcome this problem (see e.g.~\cite{mescheder2019occupancy, groueix2018, DeprelleGFKRA19}). 
Thereby, the difference between two point clouds is usually measured using the Chamfer distance (see e.g.
~\cite{ChenDYLFT20, groueix2018, ZhaoBDT19}). The popularity of the Chamfer distance stems from its ability to measure the distance from a point in one point cloud to its nearest neighbor in the other and vice versa. However, using the Chamfer metric on point cloud samples badly misguides gradient descent optimization algorithms. This is due to the fact that not all points are considered when determining the nearest neighbors of point cloud samples. Thus, distances between wrong matches are minimized with high probability. Although a local optimum can be achieved, high frequencies are usually lost. This problem is particularly serious when a point cloud is densely sampled and captures detailed structures. The most commonly used PointNet encoders \cite{ChenDYLFT20, groueix2018, ZhaoBDT19} suffer from exactly the same problem as they rely on lossy global pooling layers that can only be generalized to a limited extent. Even worse, the Chamfer distance is generally known as an unsuitable criterion for the similarity of point clouds \cite{Liu20, AchlioptasDMG18} (see Section \ref{sec:chamfer_distance}). The earth movers distance (EMD), that solves a linear assignment problem (LAP) between compared point clouds, is considered to be more appropriate \cite{Liu20, AchlioptasDMG18}. Solving a linear assignment problem, in turn, has a cubic runtime for an exact solution and even approximations are not efficient enough for training neural networks on point clouds with more than a few thousand points.

In this paper we present NeuralQAAD, a deep autodecoder for very compact representation of high resolution point clouds. NeuralQAAD is based on a specific neural network architecture that overlays multiple foldings defined on shared learned features to reconstruct even high resolution point clouds. We train NeuralQAAD through gradient descent while simultaneously solving and minimizing a newly defined quadratic assignment problem (QAP). Our main contributions are  \vspace{-\topsep}
\begin{enumerate}
  \item a new scaleable point cloud autodecoder architecture that is able to efficiently recover high resolutions, 
  \item a QAP to construct a perfect matching between point clouds, together with algorithms to efficiently approximate this QAP, 
  \item an empirical proof that the direct optimization of the Chamfer distance can disregard detailed structures,
  \item a validation that our approach outperforms the current
  state-of-the-art on high resolution point clouds, and finally
  \item a novel dataset of detailed skull CT-scans.
\end{enumerate}
\section{Related Work}
In the following, we give a brief overview of the three main areas of related work: algorithms and data structures to process point clouds using deep learning, neural network architectures for compressing point clouds, and assignment problems.

\subsection{Deep Learning for 3D Point clouds}
Various data structures for representing 3D data in the context of deep learning have been examined, including meshes \cite{BrunaZSL13}, voxels \cite{WangSLT18, MaturanaS15}, multi-view \cite{SuMKL15}, classifier space \cite{mescheder2019occupancy} and point clouds \cite{point}. Point clouds are often the raw output of 3D scanners and are characterized by irregularity, sparsity, missing neighborhood information, and permutation invariance. To cope with these challenges, many approaches convert point clouds into other representations such as discrete grids or polygonal meshes.   

Nonetheless, each data structure comes with its downsides. Voxelization of point clouds \cite{GirdharFRG16, SharmaGF16, gan3d, RoveriROG18} enables the use of convolutional neural networks but sacrifices sparsity. This leads to significantly increased memory consumption and decreased resolution. Utilizing meshes requires to reconstruct neighborhood information. This can either be done handcrafted, with the risk of wrongly introducing a bias, or learned for a particular task \cite{ChenDYLFT20}. In both cases, again, memory consumption increases significantly due to the storage of graph structures. Related to the transformation into meshes are approaches that define a convolution on point clouds \cite{TatarchenkoPKZ18, XuFXZQ18}. Embedding point clouds into a classifier space via fully connected neural networks \cite{mescheder2019occupancy, deepsdf} does not consider the advantages of point clouds as neural networks tend to smooth out the irregularity of point clouds and thereby may miss detailed structures. These methods are also not trainable out of the box as a countable finite set has to be defined within an uncountable space.

PointNet \cite{point} was the first neural network to directly work on point clouds. It applies a pointwise fully connected neural network to extract pointwise features that are combined by max-pooling to create global features. Therefore, PointNet has considerations for all previously mentioned disadvantages. There is numerous work on extending PointNet to be sensitive to local structures (see e.g.~\cite{QiYSG17, ZhouT18}). In this paper, we focus on unsupervised learning. Most of the aforementioned methods are only applicable to encoding point clouds. We deploy PointNet to produce intermediate encodings but discard it for the final compression.

\subsection{Deep Point Cloud Compression}\label{sec:deep_compression}
An autoencoder is the standard technique when it comes to compressing data using deep learning. However, recent work \cite{deepsdf} demonstrates that solely utilizing a decoder can be sufficient, at least in the 3D domain. The encoder is replaced by a trainable latent tensor. Deep generative models like generative adversarial networks (GANs) \cite{goodfellow2014generative} or variational autoencoders (VAEs) \cite{kingma2013auto} can also be applied to reduce dimensionality, but add overhead complexity and are rarely used for this purpose, although for UAEs \cite{zadeh2019variational} it may be sufficient to train only one decoding part, just like for standard autoencoders.

Most commonly, deep learning based compression methods for 3D point clouds either rely on voxelization \cite{GirdharFRG16, SharmaGF16, gan3d} or have one neuron per point in a fully connected output layer \cite{FanSG17, AchlioptasDMG18}. The latter approaches are hardly trainable for large-scale point clouds and require a huge number of parameters even for small point clouds. In contrast, FoldingNet \cite{folding} and FoldingNet++ \cite{ChenDYLFT20} rely on PointNet for encoding and use pointwise decoders that are based on the folding concept. Thereby folding describes the process of transforming a fixed input point cloud to a target point cloud by a pointwise neural network. FoldingNet++ additionally learns graph structures for localized transformations partly avoiding the problem of the smooth signal bias. AtlasNet \cite{groueix2018} is similar to the folding concept but \textit{folds} and translates multiple  2D patches to cover a target surface. AtlasNetV2 \cite{DeprelleGFKRA19} makes these patches trainable. AtlasNet and AtlasNetV2 suffer from the problem that it may be a challenging task to glue together all patches without artifacts and that they are hardly scalable. 3D Point Capsule Networks \cite{ZhaoBDT19} are also pointwise autoencoders, but use multiple latent vectors per instance to capture different basis functions while using a dynamic routing scheme \cite{SabourFH17}. All of the state-of-the-art point cloud autoencoders \cite{ChenDYLFT20, groueix2018, ZhaoBDT19} are trained by optimizing the Chamfer distance. Older approaches such as \cite{FanSG17, AchlioptasDMG18} trained on the earth mover distance, as well. In this work, we provide a \emph{surrogate loss} for minimizing the EMD even for high resolution point clouds. 

\subsection{Assignment Problems}\label{sec:ap}
In its most general form, an assignment problem consists of several agents and several tasks. The goal is to find an assignment from the agents to the tasks that minimizes some cost function. The term assignment problem is often used as a synonym for the linear assignment problem. LAPs have the same number of agents and tasks. The cost function is defined on all assignment pairs. Kuhn proposed a polynomial time algorithm, known as the Hungarian algorithm \cite{Kuhn1955Hungarian}, that can solve the LAP in quartic runtime. Later algorithms such as \cite{karp} or \cite{jonker} improved to cubic runtime. A fast and parallelizable $\epsilon$-approximation algorithm (the auction algorithm) to determine an almost optimal assignment was introduced in \cite{bertsekas} and used in \cite{FanSG17} to calculate the EMD. We use a GPU-accelerated implementation of the Auction algorithm \cite{Liu20}. Although being fast in comparison to previous algorithms, it is by no means fast enough to act as a gradient descent criterion. 

Besides the LAPs, there are nonlinear assignment problems (NAPs) which differ in the construction of the underlying cost function. For NAPs, the cost function can have more than two dimensions. In this work, we utilize a four dimensional quadratic assignment problems to model registration tasks. More precisely, we use the formulation of \cite{Beckman1957}, in which the general objective is to distribute a set of facilities to an equally large set of locations. In contrast to the LAP, there is no cost directly defined on pairs between facilities and locations. Instead, costs are modeled by a flow function between facilities and a distance function between locations. Given two facilities, the cost of their assignment is calculated by multiplying the flow between them with the distance between the locations to which they are assigned. This definition ensures that facilities that are linked by a high flow are placed close together. However, finding a solution or even a $\epsilon-$approximative solution is NP-hard \cite{burkard1984quadratic}. Hence, even for small instances, it is almost impossible to solve a QAP in a reasonable time. In this paper, we propose a new approximation algorithm for QAPs that can be efficiently executed on GPUs.

Closely connected with assignment problems are correspondence searches (see e.g.~\cite{LitanyRRBB17, Eisenberg, GroueixFKRA18}). In simplified terms, correspondences are searched between an object and a deformation of itself. In our case we are looking for \emph{correspondences} between two different objects that make it easier to \textit{fold} one into the other. A shared challenge with our task is the improvement of the Chamfer distance. \cite{LiSSPS19} introduced a structured Chamfer distance that partially avoids local optima by creating a nearest neighbor hierarchy. However, this approach is template based and thus cannot easily be transferred to our more general goal.
\begin{figure*}[h]
\centering
\includegraphics[width=0.9\textwidth]{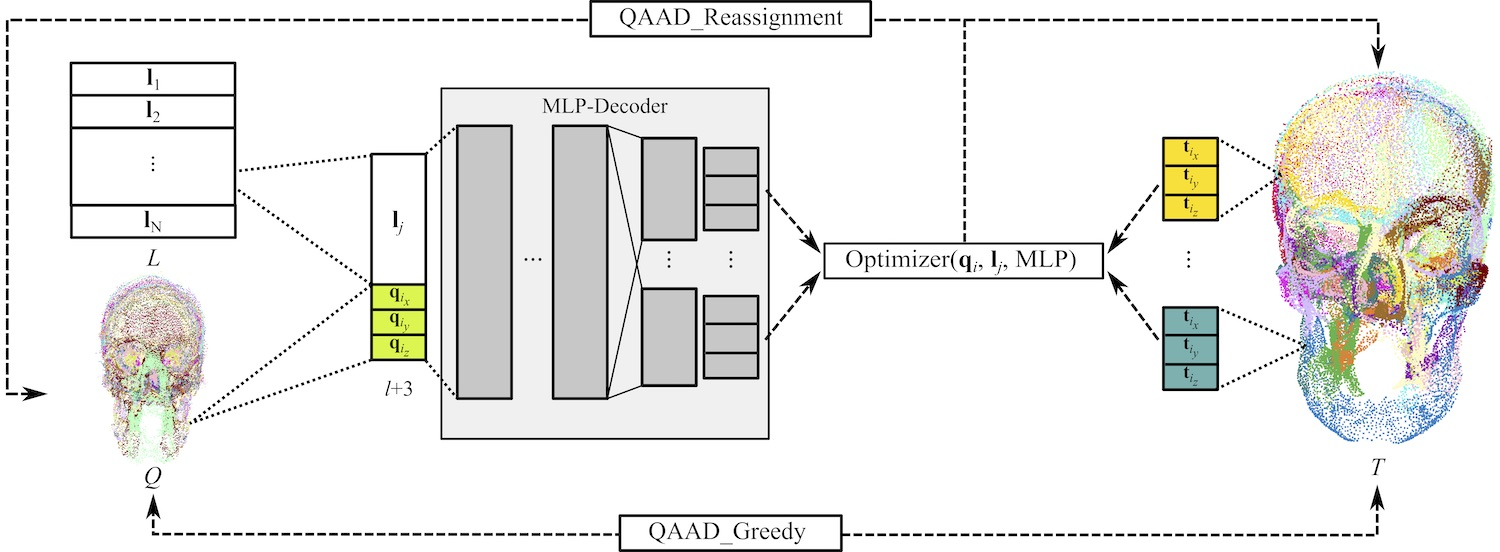}
\caption{The topology and algorithms dedicated to NeuralQAAD. The initial assignment between the input point cloud $Q$ and a target point cloud $T$ is determined by \textsc{QAAD\_Greedy}. $T$ is described by a latent vector $\mathbf{l}_j$ and stored in the lookup table $L$. A point $\mathbf{q}_i$ is processed along with $\mathbf{l}_j$ by a shared MLP to create common features. Subsequently, the common features and $\mathbf{q}_i$ are folded by a patch-specific MLP. The optimizer updates the input point $\mathbf{q}_i$, the latent code $\mathbf{l}_i$, and all MLP parameters while improving the initial assignment through \textsc{QAAD\_Reassignment}.}
\label{fig:architecture}
\end{figure*}

\section{Method}
In the following, we give a formal description of the addressed problem and derive constraints for a corresponding deep learning solution. We then construct a neural network architecture based on folding \cite{folding} that satisfies these constraints. Finally, we demonstrate how to train the system by designing a quadratic assignment problem. The complete NeuralQAAD architecture can be seen in Figure~\ref{fig:architecture}.

\subsection{Problem Formulation}\label{sec:problem}
Given a set $S = \{P_i \;|\; 1 \leq i \leq N \}$ of 3D point clouds

\begin{equation}
P_i = \left\{\mathbf{p}_{ij} \in \mathbb{R}^3 \;|\; 1 \leq j \leq M \right\}
\end{equation}
all having the same number of points, we seek for a pair of functions $f,g$ where $f:\mathbb{R}^{3\cdot M}\mapsto \mathbb{R}^{l}, l \ll M$ encodes a point cloud into a lower dimensional feature vector 
\begin{equation}
\label{featurevector}
\mathbf{l}_i = f(P_i)\in \mathbb{R}^l,
\end{equation}
and $g:\mathbb{R}^{l}\mapsto \mathbb{R}^{3\cdot M}$ decodes the feature vector $\mathbf{l}_i$ back into a point cloud
\begin{equation}
\label{decoder}
    \overline{P}_i = g(\mathbf{l}_i)
\end{equation}
such that the error
\begin{equation}
\label{eq:simpleengery}
E(f, g) =  \frac{1}{N}\sum_{i=1}^N d(P_i, \overline{P}_i),
\end{equation}
is minimal with respect to a function $d$ which measures the similarity of two point clouds. As discussed in Section~\ref{sec:deep_compression}, folding transforms equation \eqref{decoder} into
\begin{equation}
\label{eq:foldingdecoder}
    \overline{P}_i = g(\mathbf{l}_i, Q),
\end{equation}
where $Q = \left\{\mathbf{q}_{j} \;|\; 1 \leq j \leq M \right\}$ is a single fixed input point cloud. 

As point clouds possess no explicit neighborhood information and are invariant to permutations, $f$ should also have these properties. Therefore, we further divide $f$ into a pointwise local feature descriptor $\hat{f}$ transforming the pointset $P_i$ into a feature set
\begin{equation}
    \hat{F}_i = \left\{\hat{f}(\mathbf{p}_{ij}) \;|\; 1 \leq j \leq M \right\}
\end{equation}
and a permutation-invariant global feature descriptor $\Tilde{f}$ which converts the feature set into a lower dimensional feature vector 
\begin{equation}
    \Tilde{l}_i = \Tilde{f}(\hat{F}_i)\in \mathbb{R}^l.
\end{equation}
Likewise, substituting $g$ with a pointwise reconstruction function $\hat{g}$ changes \eqref{eq:foldingdecoder} into
\begin{equation}
    \hat{P}_i = \left\{\hat{\mathbf{p}}_{ij} \;|\; 1 \leq j \leq M \right\} \text{ with }\;
    \hat{\mathbf{p}}_{ij} = \hat{g}(\Tilde{l}_i, \mathbf{q}_j).
\end{equation}
Thus, instead of finding two functions $f,g$ that minimizes \eqref{eq:simpleengery}, our objective is now to find three functions $\hat{f}, \hat{g}$ and $\tilde{f}$ that minimize
\begin{equation}
\label{eq:prob}
    E(\hat{f}, \hat{g}, \Tilde{f}) = \frac{1}{N} \sum_{i=1}^N d(P_i, \hat{P}_i).
\end{equation}

Minimizing \eqref{eq:prob} using deep learning algorithms, in general, is very memory and computational intensive as e.g. a comparably small batch of \num{32} point clouds each consisting of $M=\num{65000}$  points is equal to processing a rather large batch of around \num{2653} MNIST images. Therefore the realization of $\hat{f}, \hat{g}$ and $\tilde{f}$ through deep neural networks requires computationally efficient and memory-optimized methods as presented in the following. 

\subsection{EM-kD}\label{sec:emkd}
The measure $d$ for the similarity of point clouds is typically chosen to be either the Chamfer distance \cite{groueix2018,ZhaoBDT19} 
\begin{equation}
    d_C(P, \hat{P}) = \sum_{\mathbf{p}\in P} \min_{\hat{\mathbf{p}} \in \hat{P}} \left\Vert \mathbf{p} - \hat{\mathbf{p}} \right\Vert_2 +
    \sum_{\hat{\mathbf{p}}\in\hat{P}} \min_{\mathbf{p} \in P} \left\Vert \mathbf{p} - \hat{\mathbf{p}} \right\Vert_2
\end{equation}
or the augmented Chamfer distance \cite{ChenDYLFT20} 
\begin{equation}
    d_A(P, \hat{P}) = \max \left\{ 
    \sum_{\mathbf{p}\in P} \min_{\hat{\mathbf{p}} \in \hat{P}} \left\Vert \mathbf{p} - \hat{\mathbf{p}} \right\Vert_2, 
    \sum_{\hat{\mathbf{p}}\in\hat{P}} \min_{\mathbf{p} \in P} \left\Vert \mathbf{p} - \hat{\mathbf{p}} \right\Vert_2
    \right\}.
\end{equation}
Besides previous detected weaknesses  \cite{Liu20,AchlioptasDMG18}, we further show in Section~\ref{sec:chamfer_distance} that $d_C$ and $d_A$ are additionally vulnerable to an optimizer getting stuck in weak local minima if high resolution point clouds with dense sampling rates are processed. 
 
Compared to the Chamfer distance, the Earth Mover Distance (EMD)
 \begin{equation}
    d_{EMD}(P, \hat{P}) = \min_{\phi:P \rightarrow \hat{P}}\sum_{\mathbf{p}\in P} \left\Vert \mathbf{p} - \mathbf{\phi(p)}\right\Vert_2 
\end{equation}
is considered to be more representative for visual varieties \cite{Liu20,AchlioptasDMG18}. 
Since the use of the EMD leads to intractable runtimes even for moderately large point clouds, we introduce the EM-kD, a divide-and-conquer approximation of the true EMD. 

Algorithm \ref{algo:emkd} details the calculation of the EM-kD. First, both the source and target point clouds are spatially partitioned through constructing k-d trees of the same size. The resulting subspaces can be canonically matched i.e. by pairing subspaces with the same position in the respective k-d trees. We then proceed by running the \emph{auction algorithm} \cite{bertsekas} on each subspace pair and finally average over all calculated distances. 

 \algrenewcommand\algorithmicindent{0.9em}
\begin{algorithm}[htp]
\caption{Approximation of the Earth Movers Distance between a \emph{Source} and a \emph{Target} point cloud by partitioning both into $2^{depth}$ many subspaces and utilizing the auction algorithm for each canonical pair of \emph{Source} and \emph{Target} subspaces.}
\label{algo:emkd}
\begin{algorithmic}
\fontsize{8}{8}
\Procedure{EM-kD}{\texttt{Source}, \texttt{Target}, \texttt{depth}} 
    \State // Build up k-d trees
    \State \texttt{SLeafs $\leftmapsto$ kdTree(Source, \texttt{depth})}
    \State \texttt{TLeafs $\leftmapsto$ kdTree(Target, \texttt{depth})}
    \texttt{\\}
    \State // Match each subspace pairs
    \State \texttt{sumDistance $\leftmapsto$ 0}
    \For{\texttt{i} = $1, 2, \ldots, 2^{depth}$ } 
    \State // MatchIDs are indices in the TargetTree subspace
    \State \texttt{MatchIDs,\!\!\! Dists}  $\leftmapsto$ \texttt{auction(SLeafs[i],\!\!\! TLeafs[i])}
    \State // Add the average distance of the subspace
    \State \texttt{sumDistance $\leftmapsto$ sumDistance + avg(Dists)}
    \EndFor 
    \State
    \State // Return the average distance of all subspaces 
    \State \Return \texttt{sumDistance / $2^{\texttt{depth}}$}
\EndProcedure
\end{algorithmic}
\end{algorithm}

While being a reasonable upper bound of the true EMD, EM-kD is greatly scaleable as the subtask size of the auction algorithm can be controlled through the depth of the k-d trees. We use EM-kD for evaluating NeuralQAAD. In Section~\ref{sec:training} we present a training procedure to implicitly optimizing the EM-kD and hence the EMD. 

\subsection{Network Architecture}
In the following, we discuss the three key components for a successful implementation of a scalable autodecoder for point clouds based on a deep neural network. Since an essential part of our new neural network-based autodecoder is the efficient solution of a \emph{quadratic assignment problem}, we call it NeuralQAAD.

First, NeuralQAAD follows the idea of folding. In contrast to previous work that folds a fixed \cite{folding} or trainable \cite{DeprelleGFKRA19} point cloud by uniformly sampling a specific manifold such as a square, we choose to deform a random sample of the target data set as this is very likely to be closer to any other sample than an arbitrary primitive geometry and thus better suited for initialization. For a dataset with strong low frequency changes, clustering of multiple random examples may be even more advantageous. We leave this for future research as this turned out to be not beneficial in our experiments.

Second, we found that splitting the input point cloud into $K$ many patches that are independently folded \cite{DeprelleGFKRA19}, i.e. $K$ independent multilayer perceptrons (MLPs), is essential but also makes highly inefficient use of resources. Each patch has its own parameter set and hence its own set of extracted features. Increasing the number of patches and as a result the number of MLPs quickly restricts the depth of those. We observe that weight sharing across the first layers of each MLP does not decrease the performance. Quite the contrary, with the same number of parameters we can design a deeper and/or wider architecture. Likewise, the choice of sharing low-level features can also be used to scale the number of patches.

Third, as discussed in Section~\ref{sec:problem}, a batch of large point clouds may not be processed at once. To cope with this problem, NeuralQAAD only \textit{folds} a sample from each point cloud of a batch during a single forward pass. Since the latent code $\tilde{l}_i$ of an instance depends on all its related points, the encoding function $\Tilde{f}$ needs to be adapted. We realize $\Tilde{f}$ as a trainable lookup table and thus renounce the explicit calculation of the pointwise function $\hat{f}$. Besides our motivation for autodecoders to minimize memory usage, \cite{deepsdf} suspect that this topology uses computing resources more effectively. In contrast to \cite{mescheder2019occupancy}, we could not find any evidence that conditioning through conditional batch normalization \cite{VriesSMLPC17} is beneficial.

Considering the three constituents just discussed, the problem of finding functions $\hat{f},\hat{g}$ and $\Tilde{f}$ that minimize \eqref{eq:prob} can now be rephrased into the problem of training multiple multilayer perceptrons $MLP_k$, one for each input patch $Q_k$, and a lookup table $L\in\mathbb{R}^{N\times l}$ such that 
\begin{equation}
\label{eq:finprob}
    E(MLP, L, Q) = \frac{1}{N}\sum_{i=1}^N d(P_i, \hat{P}_i),
\end{equation}
is minimized, where the elements of $\hat{P}_i$ are given as
\begin{equation}
\hat{\mathbf{p}}_{i,j} = \{MLP_k(\mathbf{l}_i, \mathbf{q}_{k,j})\}_{k \in K}.
\end{equation}
Without loss of generality we do not explicitly state the split into various patches for simplification of notation in the following.

\subsection{Training}\label{sec:training}
Optimizing NeuralQAAD only on subsets of the input point cloud $Q$ directly affects the construction of the loss function. The common pattern for calculating a distance between points clouds is similar to a registration task that can be split into the two steps
\begin{enumerate}
\item Identifying a suitable matching between both point clouds.
\item Determine the actual deformation of one point cloud to the other. 
\end{enumerate}
Gradient descent is only done for the second step whereas the first step characterizes how appropriate a metric is. Besides the Chamfer distance not being reliable and the EMD not being tractable a general consideration weights in. The matching that steers the folding process has to comply with the predominant bias of neural networks, which is smoothing. Therefore, we choose a matching that realizes the \emph{quadratic assignment problem} described next. 

Given an input point cloud $Q$, a target point cloud $P$, a weight function $w(\mathbf{p}, \hat{\mathbf{p}}) = \left\Vert \mathbf{p} - \hat{\mathbf{p}} \right\Vert_2$, and a flow function $f(\mathbf{q}, \hat{\mathbf{q}}) =\nicefrac{1}{w(\mathbf{q}, \hat{\mathbf{q}})}$ we want to find a bijective mapping $A:Q \mapsto P$ such that
\begin{equation}
    E(A)=\sum_{\mathbf{q}, \hat{\mathbf{q}} \in Q} f(\mathbf{q},\hat{\mathbf{q}}) \cdot w(A(\mathbf{q}),A(\hat{\mathbf{q}})) 
\end{equation}
is minimized. Intuitively speaking, this means that by solving this quadratic assignment problem, we assure that points close to each other in $Q$ are matched with points close to each other in $P$. Note that LAPs do not exhibit this property but solving the QAP also allows us to optimize the EMD. 

\algrenewcommand\algorithmicindent{0.9em}
\begin{algorithm}[htp]
\caption{Calculate an initial QAP solution between a \emph{Source} and a \emph{Target} by applying the matches of the EM-kD in the subspaces \emph{SLeafs} and \emph{TLeafs}. The algorithm assumes that both point clouds are stored as $\mathbb{R}^{n \times 3}$ matrices and points are matched if they have the same row index.}
\label{algo:qaadGreedy}
\begin{algorithmic}
\fontsize{8}{8}
\Procedure{QAAD\_Greedy}{\texttt{Source}, \texttt{Target}, \texttt{depth}} 
    \State // Build up k-d trees
    \State \texttt{SLeafs $\leftmapsto$ kdTree(Source, \texttt{depth})}
    \State \texttt{TLeafs $\leftmapsto$ kdTree(Target, \texttt{depth})}
    \State
    \State // Match each subspace pairs
    \For{\texttt{i} = $1, 2, \ldots, 2^{depth}$ } 
    \State // MatchIDs are indices in the TargetTree subspace
    \State \texttt{MatchIDs,\!\!\! Dists $\!\!\!\!\leftmapsto\!\!\!\!$ auction(SLeafs[i],\!\!\! TLeafs[i])}
    \State \texttt{SourceIDs $\leftmapsto$ range(1, size(MatchIDs))}
    \State
    \State // Auction algorithm only produces almost perfect bijection
    \State // If repeatedly assigned, use smallest distance
    \State \texttt{ArgIDs $\leftmapsto$ argsort(Dists[Swap])}
    \State \texttt{SourceIDs $\leftmapsto$ SourceIDs[ArgIDs]}
    \State \texttt{MatchIDs $\leftmapsto$ MatchIDs[ArgIDs]}
    \State
    \State // After sorting simply consider first occurrence of MatchIDs
    \State // Randomly match remaining points
    \State \texttt{FirstIDs $\leftmapsto$ firstOccurence(MatchIDs)}
    \vspace{0.1cm}
    \State \texttt{SourceIDs $\leftmapsto$ SourceIDs[FirstIDs]}
    \State \texttt{RemainingSourceIDs $\leftmapsto$ not SourceIDs}
    \vspace{0.1cm}
    \State \texttt{MatchIDs $\leftmapsto$ MatchIDS[FirstIDs]}  
    \State \texttt{RemainingMatchIDs $\leftmapsto$ not MatchIDs}
    \vspace{0.1cm}
    \State \texttt{SourceIDs $\!\!\!\!\leftmapsto\!\!\!\!$ cat(SourceIDs,\!\!\!\! RemainingSourceIDs)}  
    \State \texttt{MatchIDs $\leftmapsto$ cat(MatchIDs, RemainingMatchIDs)}  
    \State
    \State // Conduct beneficial reassignments
    \State \texttt{Tmp $\leftmapsto$ Target[SourceIDs]}
    \State \texttt{Target[SourceIDs] $\leftmapsto$ Target[MatchIDs]}
    \State \texttt{Target[MatchIDs] $\leftmapsto$ Tmp}
    \EndFor 
\EndProcedure
\end{algorithmic}
\end{algorithm}
\begin{algorithm}[t]
     \floatname{algorithm}{Algorithm}
\begin{algorithmic}[0]
\fontsize{8}{8}
\caption{Refining the initial QAP solution for a query of points \emph{SQueryIDs} of the input point cloud \emph{Source}. At this, the image of the query given a latent code \texttt{code} is constructed by the current state of the NeuralQAAD MLP \emph{model} and evaluated against the target point cloud \emph{Target}. However, for efficiency only a \emph{TQueryIDs} of \emph{Target} is considered.}
\label{algo:qaadReassignment}
\Procedure{QAAD\_Reassignment}{\texttt{Source}, \texttt{Target}, \texttt{code},\newline 
\hspace*{44mm}\texttt{SQueryIDs}, \texttt{TQueryIDs} } 
    \State // Predict query and determine nearest neighbors in target
    \State \texttt{Predict $\leftmapsto$ model(Source[SQueryIDs], code)}
    \State \texttt{NnIDs,\!\!\! Dists $\!\!\!\leftmapsto\!\!\!$ nn(Predict,\!\!\! Target[TQueryIDs])}
    \State
    \State // Predict cube points that are matched to nearest neighbors
    \State \texttt{PredictNn $\leftmapsto$ model(Source[NnIDS], code)}
    \State
    \State// Check current loss
    \State \texttt{LossBefore $\!\!\!\leftmapsto\!\!\!$ norm(Predict,\!\!\! Target[SQueryIDs])}
    \State \texttt{LossBeforeNn $\!\!\!\!\leftmapsto\!\!\!\!$ norm(PredictNn,\!\! Target[NnIDs])}
    \State \texttt{LossBefore $\leftmapsto$ LossBefore + LossBeforeNn} 
    \State
    \State// Check loss after hypothetical reassignment
    \State \texttt{LossAfter $\leftmapsto$ norm(Predict, Target[NnIDs])}
    \State \texttt{LossAfterNn $\!\!\!\!\leftmapsto\!\!\!\!$ norm(PredictNn,\!\!\! Target[SQueryIDs])}
    \State \texttt{LossAfter $\leftmapsto$ LossAfter + LossAfterNn} 
    \State
    \State // Decide where reassignment is beneficial
    \State \texttt{Swap $\leftmapsto$ LossAfter < LossBefore}  
    \State \texttt{SQueryIDs $\leftmapsto$ SQueryIDs[Swap]}  
    \State \texttt{NnIDs $\leftmapsto$ NnIDs[Swap]}  
    \State
    \State // If repeatedly assigned, use smallest distance
    \State \texttt{ArgIDs $\!\!\!\leftmapsto\!\!\!$ argsort((LossAfter \!\!-\!\! LossBefore)[Swap])}
    \State \texttt{SQueryIDs $\leftmapsto$ SQueryIDs[ArgIDs]}
    \State \texttt{NnIDs $\leftmapsto$ NnIDs[ArgIDs]}
    \State // After sorting simply consider first occurrence of NnIDs
    \State \texttt{FirstIDs $\leftmapsto$ firstOccurence(NnIDs)}
    \State \texttt{SQueryIDs $\leftmapsto$ SQueryIDs[FirstIDs]}
    \State \texttt{NnIDs $\leftmapsto$ NnIDS[FirstIDs]}
    \State
    \State // If a query point wants to swap its matched target,
    \State // the other source point must be the exclusive partner
    \State \texttt{ExclusiveIDs $\leftmapsto$ where QueryIDs notIn NnIDs}
    \State \texttt{SQueryIDs $\leftmapsto$ SQueryIDs[ExclusiveIDs]}
    \State \texttt{NnIDs $\leftmapsto$ NnIDs[ExclusiveIDs]}
    \State
    \State // Conduct beneficial reassignments
    \State \texttt{Tmp $\leftmapsto$ Target[SQueryIDs]}
    \State \texttt{Target[SQueryIDs] $\leftmapsto$ Target[NnIDs]}
    \State \texttt{Target[NnIDs] $\leftmapsto$ Tmp}
\EndProcedure
\end{algorithmic}
\end{algorithm}
On first sight, QAPs seem to be harder to solve than LAPs. However, we propose a two-step approach to find a sufficiently accurate QAP matching that allows for training NeuralQAAD with only a small overhead to previous approaches. The first step is a greedy algorithm called \textsc{QAAD\_Greedy} that generates an initial matching before gradient-descent. The second step is a dynamic reassignment algorithm called \textsc{QAAD\_Reassignment}, that monotonically improves equation \eqref{eq:finprob}. During our experiments it became evident that neither of the two steps alone leads to an accurate solution. NeuralQAAD can also be trained with the AtlasNetV2 \cite{DeprelleGFKRA19} training procedure which still results in a significant performance improvement but does not produce any overhead at all (see Section \ref{sec:exp}).

\textsc{QAAD\_Greedy} (see Algorithm~\ref{algo:qaadGreedy}) is conceptually similar to EM-kD. Instead of using the calculated distance between the input and a target point cloud the determined LAP matches are used as the initial QAP solution. To establish a perfect matching that is not guaranteed by the auction algorithm, the input point that is closest to a repeatedly chosen target point is matched. Any remaining points are randomly matched. Like for the EM-kD, all operations of \textsc{QAAD\_Greedy} can be efficiently parallelized on a GPU. Even for the biggest tested datasets \textsc{QAAD\_Greedy} run only minutes. As \textsc{QAAD\_Greedy} is only run once, the runtime is negligible considering the common training time of neural networks.

Since the input point cloud is trainable, \textsc{QAAD\_Greedy} may make assignment errors at subspace borders, and the initial matching is not regularized for smoothness, \textsc{QAAD\_Reassignment} (see Algorithm~\ref{algo:qaadReassignment}) conducts reassignments during gradient descent that ease the training of NeuralQAAD. In contrast to the initial matching, \textsc{QAAD\_Re\-as\-sign\-ment} uses the image of the input points. At this, given a latent code, a sample of input points is predicted and the nearest neighbor from a sample of the target points is found for each prediction. This can be efficiently done in parallel on a CPU using k-d trees or other suitable data structures. Newer GPU architectures that refrain from warp-based computation also allow for efficient GPU implementations. The k-d trees are build and queried with only a few samples per training step and, hence add little overhead. Subsequently, the input points that are actually assigned to the nearest neighbors are also predicted. Reassignment is based on comparing the summed loss of both predictions under the actual matching with the summed loss of both predictions after a hypothetical pointwise swap. In case the latter loss is smaller, a reassignment is conducted. Hence, each reassignment is guaranteed to reduce the error defined by equation \eqref{eq:finprob}. Just like for \textsc{QAAD\_Greedy}, a perfect matching must be maintained. However, random sampling is not applicable here as this would undermine the intention of \textsc{QAAD\_Re\-as\-sign\-ment}. Instead, if two input points want to swap for the same target point, only the swap with the strongest decline in distance is conducted. Further, an input point that wants to swap for a target can not be the input for the target point of another swap.

\begin{algorithm}
     \floatname{algorithm}{Algorithm}
\begin{algorithmic}[0]
\fontsize{8}{8}
\caption{Training of the individual NeuralQAAD components on a target dataset $Dataset$, the initially source point cloud $Source$, the pretrained folding MLP $model$, and a latent code table $CodeTabel$. The training is done for $numEpochs$ epochs in batches of size $batchSize$ with $sampSize$ many points per instance.}
\label{algo:training}
\Procedure{Training}{\texttt{CodeTable}, \texttt{Source}, \texttt{Dataset}, \texttt{depth}\newline
\hspace*{26mm}\texttt{numEpochs}, \texttt{batchSize}, \texttt{sampSize}}
    \State \texttt{numBatches $\leftmapsto$ |Dataset|/batchSize}
    \State // Generate initial matching for each target point cloud
    \For{\texttt{i} = $1, 2, \ldots, |\texttt{Dataset}|$}
    \State \textsc{QAAD\_Greedy}\texttt{(Source, Dataset[i], depth)}
    \EndFor
    \State
    \For{\texttt{i} = $1, 2, \ldots, \texttt{numEpochs}$}
        \For{\texttt{j} = $1, 2, \ldots, \texttt{numBatches}$}
        \State // Get a batch of target point clouds and \State // corresponding latent codes
        \State \texttt{Target $\leftmapsto$ getTargetBatch(j, Dataset)}
        \State \texttt{Codes $\leftmapsto$ getCodeBatch(j, CodeTable)}
        \State
        \State // Sample and predict points
        \State \texttt{SQueryIDs $\!\!\!\leftmapsto\!\!\!$ uniform((1,|Source|),\!\!\! sampSize)}
        \State \texttt{Predict} $\leftmapsto$ \texttt{model(Source[SQueryIDs], Codes)}
        \State
        \State // Conduct reassignment
        \State \texttt{TQueryIDs $\!\!\!\leftmapsto\!\!\!$ uniform((1,|Source|),\!\!\! sampSize)}
        \For{\texttt{k} = $1, 2, \ldots,\texttt{batchSize}$} \textbf{in parallel}
        \State \begin{varwidth}[t]{\linewidth}
        \textsc{QAAD\_Reassignment}\texttt{Source}, \texttt{Target[k]},\par 
        \hspace*{32mm}\texttt{SQueryIDs}, \texttt{TQueryIDs},\par 
        \hspace*{32mm}\texttt{Codes[k]}
        \end{varwidth}
        \EndFor
        \State
        \State // Calculate loss and perform an optimization step
        \State \texttt{optimize(loss(Predict,\!\!\! Target[,SQueryIDs]))}
        \EndFor
    \EndFor
\EndProcedure
\end{algorithmic}
\end{algorithm}

Algorithm~\ref{algo:training} shows the overall training algorithm of NeuralQAAD by combining \textsc{QAAD\_Reassignment}, sampling,  and \textsc{QAAD\_Greedy}. Before gradient descent, the initial matching for each training instance is found by \textsc{QAAD\_Greedy}. During gradient descent, a sample of input points is drawn per batch instance and training step, which is then processed by NeuralQAAD. Before the loss is backpropagated, a potential reassignment is tested by \textsc{QAAD\_Reassignment} with a random sample of target points and applied if beneficial. 

In general, the loss can be arbitrarily chosen. Somewhat surprisingly, we found that using the augmented Chamfer loss can be beneficial. Due to the initial matching and reassignments, the augmented Chamfer distance is mostly equivalent to simply calculating the euclidean distance. However, in the rather rare case in which mismatches are sampled, it accelerates reassignment.  We enforce the sampling procedure to output the same number of points for each patch. This allows us to efficiently implement NeuralQAAD as a grouped convolution. In contrast to previous implementations that sequentially run each MLP per patch, we can make use of all GPU resources and achieve a significant speed-up.
    
\section{Experiments}\label{sec:exp}
In this section, we validate NeuralQAAD as well as its underlying assumptions. We start with an overview of the datasets used and some implementation details. Next, we demonstrate that directly optimizing the standard Chamfer loss or the augmented Chamfer loss through gradient descent suffers from weak local-minima. Finally, we show that NeuralQAAD offers significantly better performance for high resolution points clouds. At this, we compare against the AtlasNetV2 deformation architecture \cite{DeprelleGFKRA19}, the only other state-of-the-art differentiable approach that is out-of-the-box applicable to point clouds with more than a few thousand points.

\subsection{Datasets}
Our experiments are based on the three datasets Skulls, COMA \cite{COMA}, and D-Faust \cite{Dfaust}. Skulls is an artificially generated point cloud dataset constructed using a multilinear model \cite{AchenbachBGHSSB18} that is based on a skull template fitted to \num{42} CT scans of human skulls \cite{GietzenBAHSBSS2019}. The dataset consists of \num{4096} point clouds. Each point cloud is composed of \num{65536} points and includes fine-grained parts such as the nasal bones or teeth. Different from the other datasets it not only consists of surfaces but also of volumetric structures. We will make the dataset public as a new high resolution point cloud test dataset. COMA captures \num{20466} scans of extreme expressions from 12 different subjects that have been collected by a multi-camera stereo setup. D-Faust is a 4D full body dataset that includes \num{40000} multi-camera scans of human subjects in motion. We use about \num{6576} scans that contain more than \num{110000} points and about \num{4729} scans that contain more than \num{130000} from COMA and D-Faust respectively.

\begin{figure*}[ht!]
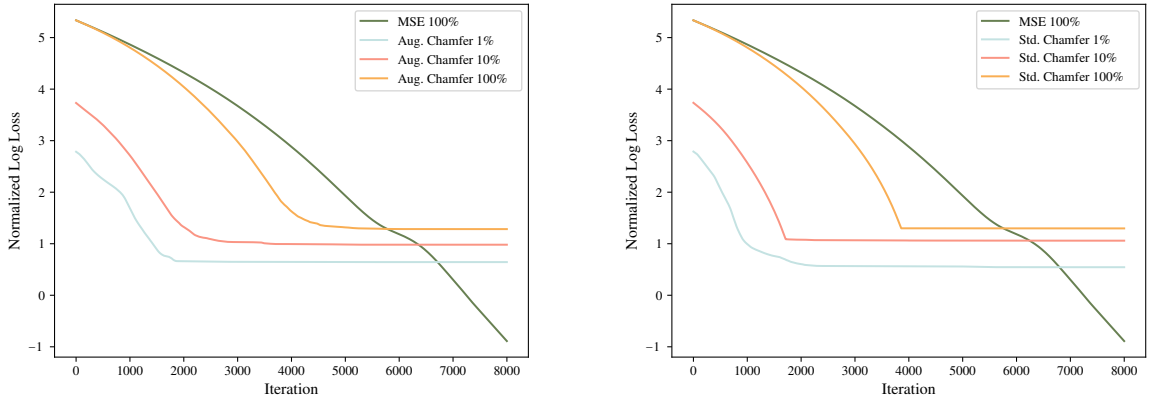

    \centering
    \scalebox{0.5}{\input{fig/chamfer_comparisson_aug.pgf}}
    \scalebox{0.5}{\input{fig/chamfer_comparisson_std.pgf}}
    \caption{The normalized logarithm of the augmented Chamfer distance between a trainable tensor and a Skulls instance over the course of gradient descent. We conduct this experiment with 100\%, 10\%, and 1\% of the targeted points. We either directly optimize the augmented Chamfer distance (see (a)) or utilize the standard Chamfer distance (see (b)) as a proxy. Neither is able to capture more detailed structures in case the sampling rate is improved. In comparison,  minimizing the MSE as a proxy, calculated between randomly assigned source-target point pairs, does not suffer from this problem.}
    \label{fig:graph_chamfer}
\end{figure*}

\subsection{Implementation Details}
For all our experiments we implement the NeuralQAAD folding operation with six fully connected layers of size 256 using PyTorch. The first four layers are shared across all patches (see Figure~\ref{fig:architecture} for an architecture overview). To guarantee the same number of parameters for the decoder while allowing AtlasNetV2 to be sufficiently deep and wide we construct the folding operation of AtlasNetV2 with four layers of size 128 and no sharing and adapt the number of patches if needed.  SELU \cite{KlambauerUMH17} is applied as a nonlinearity to every but the output layer for both NeuralQAAD and AtlasNetV2. Adam \cite{KingmaB14} is used as the optimizer and initialized with a learning rate of $0.001$. We sample 2048 points for Skulls and 4096 points for COMA as well as D-Faust per instance and training step, resulting in low video RAM consumption. AtlasNetV2 is trained until we can no longer observe any significant improvement, which was the case for all data sets after 150 epochs at the latest. To show that our architecture is also superior without our novel QAP training scheme we also train NeuralQAAD for 150 epochs with the AtlasNetV2 training scheme. Essentially, this means minimizing the augmented Chamfer loss on point cloud samples. For this purpose, we deploy a PointNet encoder, too, which we discard as soon as our QAP training scheme is utilized. Nevertheless, the PointNet embeddings are reused as initialization of the lookup table. All experiments are conducted on TITAN RTX GPUs and a AMD Ryzen™ Threadripper™ 3970X CPU. In comparison, the experiments of AtlasNetV2 have a noticeably longer runtime than those of NeuralQAAD since we use the released sequential implementation. For all experiments, we use the same hyperparameters for the auction algorithm i.e. 100 iterations with an epsilon of 1. The EM-kD is calculated with a subspace size of 1024.

\subsection{Optimizing the Chamfer distance}\label{sec:chamfer_distance}
The augmented as well as the standard Chamfer distance can be optimized either directly or via a proxy. Obviously, both Chamfer distances reach their global minima if the compared point clouds are identical. However, by construction, Chamfer distances underrate high frequencies. In the following, we demonstrate that direct optimization of the Chamfer distance using gradient descent can lead to a loss of detailed structures. We also show that this effect is amplified for high resolution point clouds with dense sampling rates.

In our experimental setup, we train an offset tensor that is added to a point cloud cube to equal a target point cloud. For each point in the input point cloud, the distance to its nearest neighbor is set to 1. We choose a random instance from the Skulls dataset as the target and train the offset through gradient descent. The experiment is conducted with all, 10\%, and 1\% of the points where subsets are uniformly drawn without replacement. To compare the different settings, we state the logarithm of the augmented Chamfer distance normalized by an approximation of the sampling rate of the target point cloud. The normalization factor of a point cloud $P$ is defined as
\begin{equation}
T(P) = \frac{1}{|P|\cdot k}\sum_{\mathbf{p} \in P}\;
\sum_{\hat{\mathbf{p}} \in kNN(\mathbf{p})} \hat{\mathbf{p}},
\end{equation}
where $kNN(\mathbf{p})$ denotes the set of the $k$ nearest neighbors of point $\mathbf{p}$. In all of our experiments we set $k=5$.

Figure~\ref{fig:graph_chamfer} shows the development of the reconstruction loss during training. In Figure~\ref{fig:graph_chamfer} right, the augmented Chamfer distance is optimized directly, whereas in Figure~\ref{fig:graph_chamfer} left the standard Chamfer distance is used as a proxy. Both indicate that a denser sampling misguides the training procedure more strongly, moves the solution further away from a perfect matching, and culminates in a weaker local optimum. As a consequence, detailed structures may be missed. The Figure~\ref{fig:graph_chamfer} also shows the well-conditioned case in which a random perfect matching between the two point clouds is given as a sanity check.  In this case, the average mean squared error (MSE) of all assignments acts as a proxy for the Chamfer distances. Our results show that omitting the direct optimization of a Chamfer distance is advantageous if a suitable perfect matching can be found. This holds in particular for high resolution point clouds with a dense sampling rate.

\subsection{Reconstruction of High Resolution Point Clouds}
\textbf{Skulls:} The Skulls dataset contains point clouds that can be linearly reduced to 42 latent variables using principal component analysis. In addition, we nonlinearly compress each skull to a latent vector of size 10 in our experiments. Although this seems to be an easy task at first sight, results of AtlasNetV2 demonstrate the contrary.
\begin{figure*}[t]
\begin{tabular}{ >{\centering\arraybackslash}m{0.31\linewidth} >{\centering\arraybackslash}m{0.31\linewidth} >{\centering\arraybackslash}m{0.31\linewidth} }
\toprule
                \footnotesize NeuralQAAD & \footnotesize Original & \footnotesize AtlasNetV2 \\\midrule
\footnotesize \includegraphics[height=55mm]{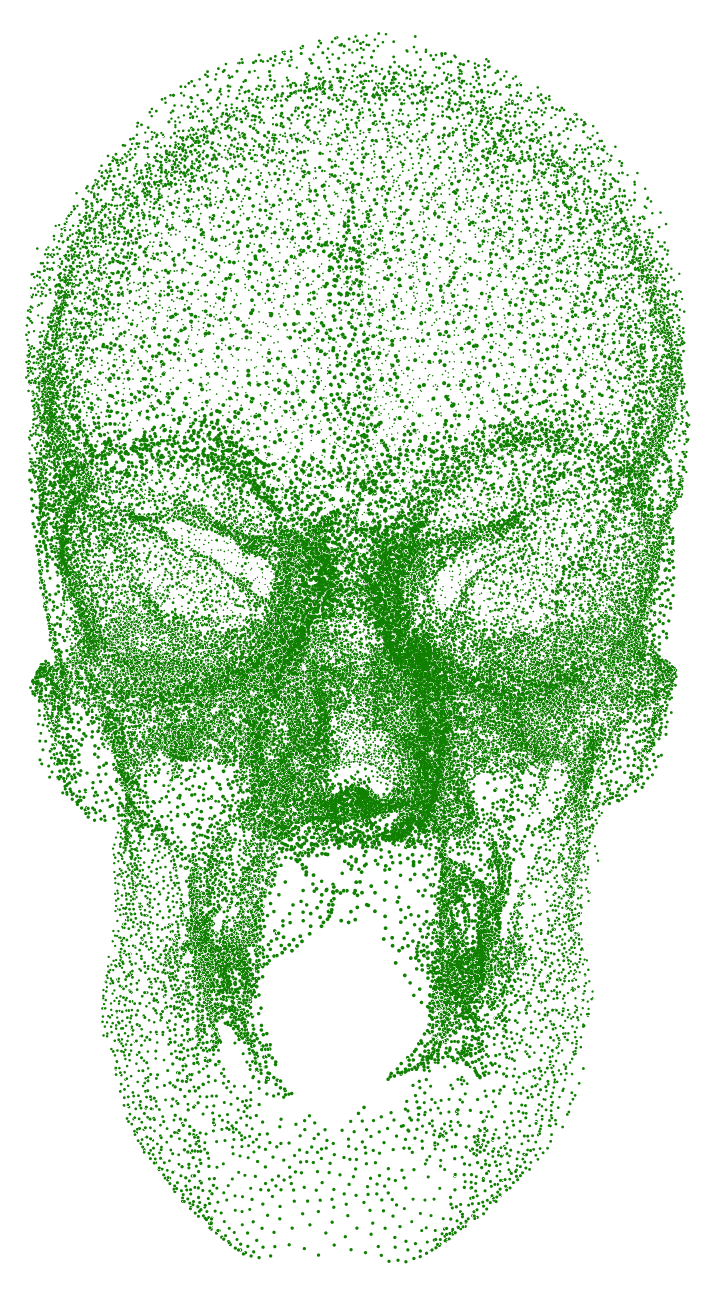}  &
                  \includegraphics[height=55mm]{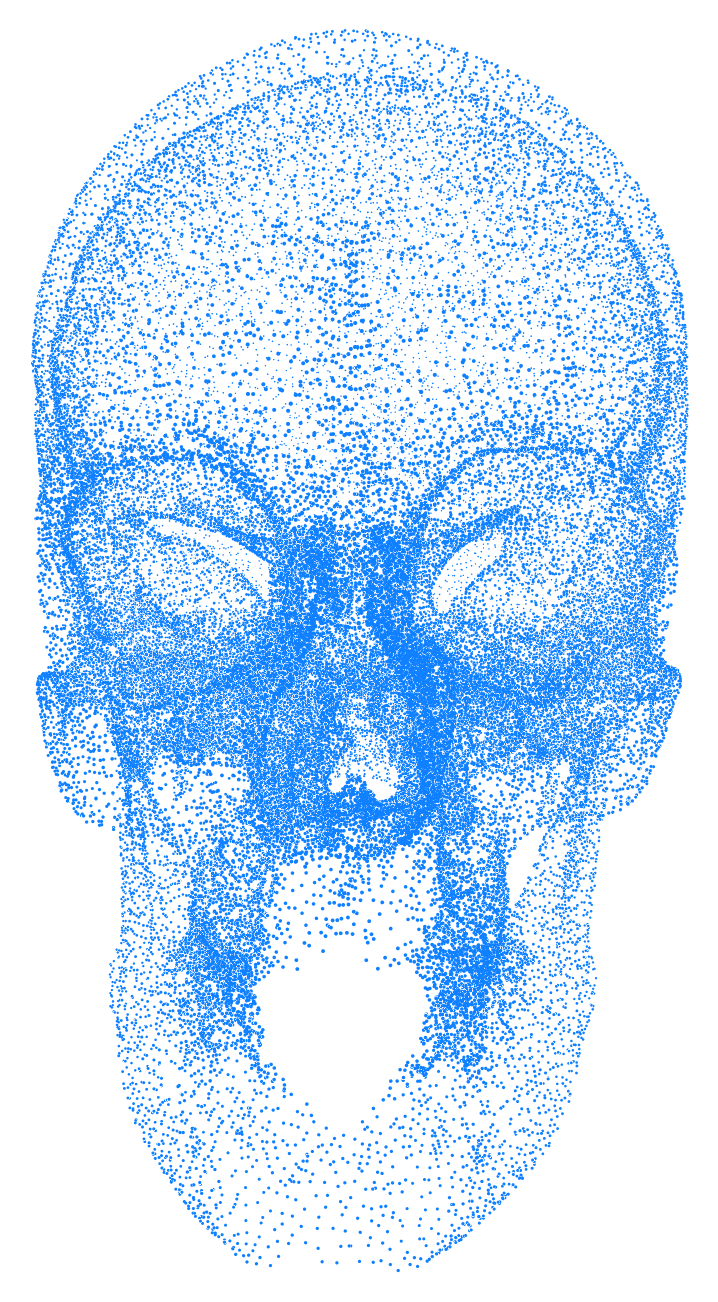}   &
                  \includegraphics[height=55mm]{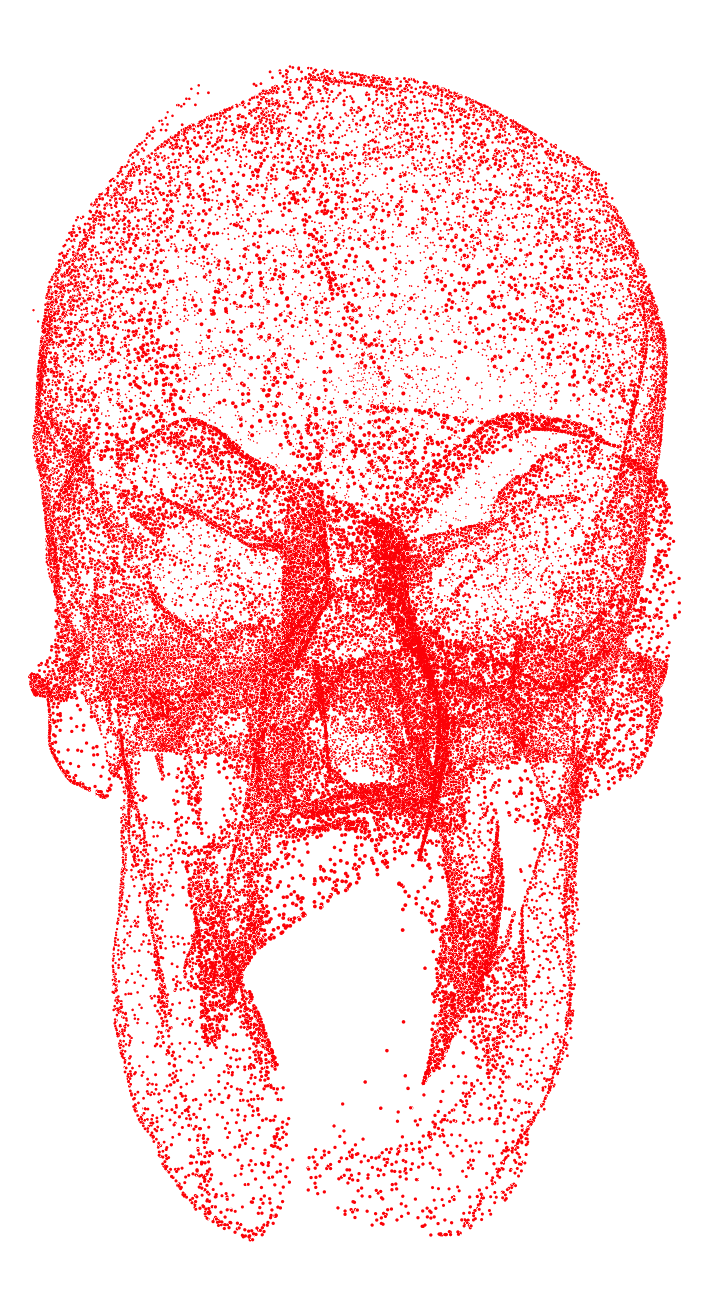}\\

\footnotesize \includegraphics[height=55mm]{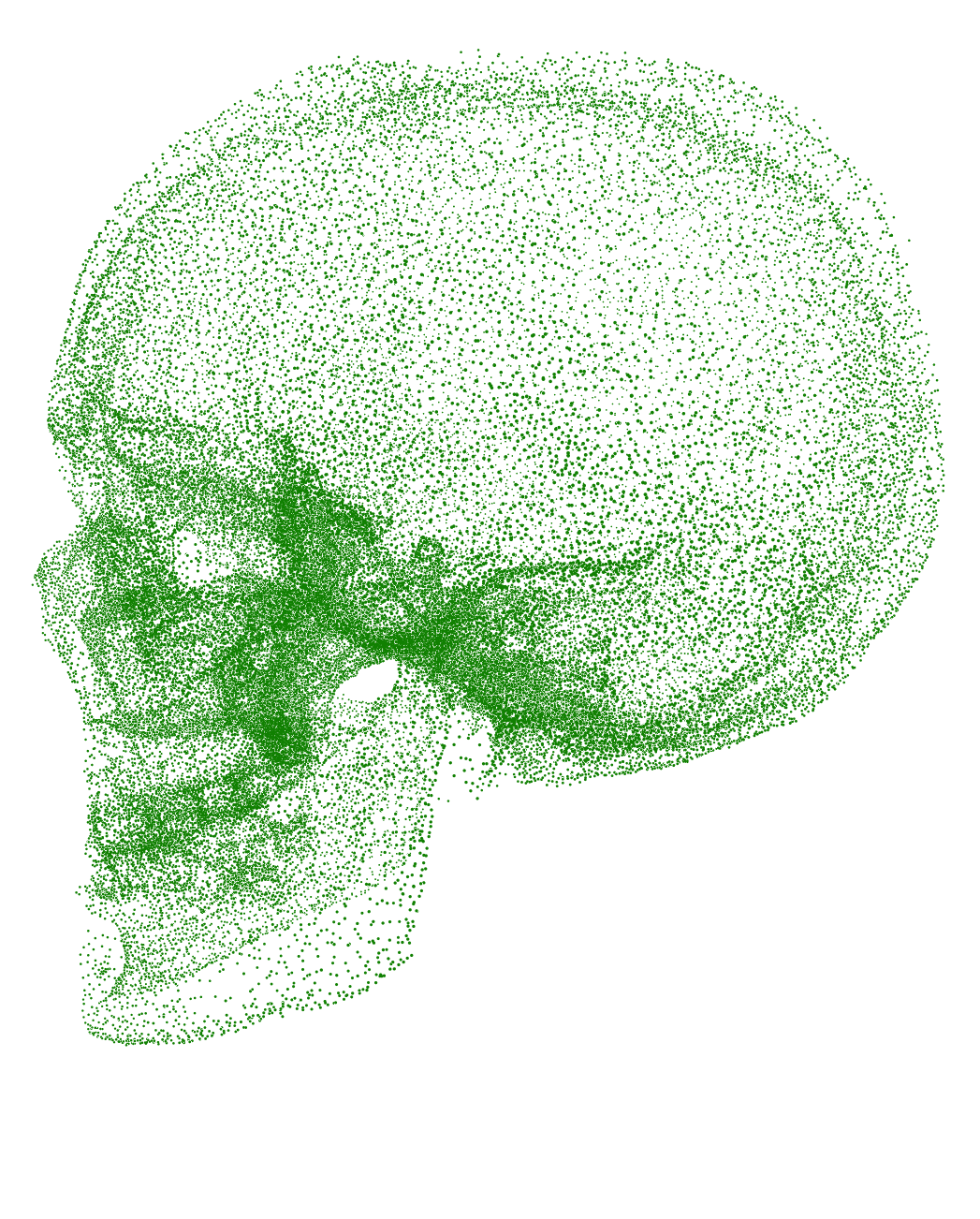}      &
                  \includegraphics[height=55mm]{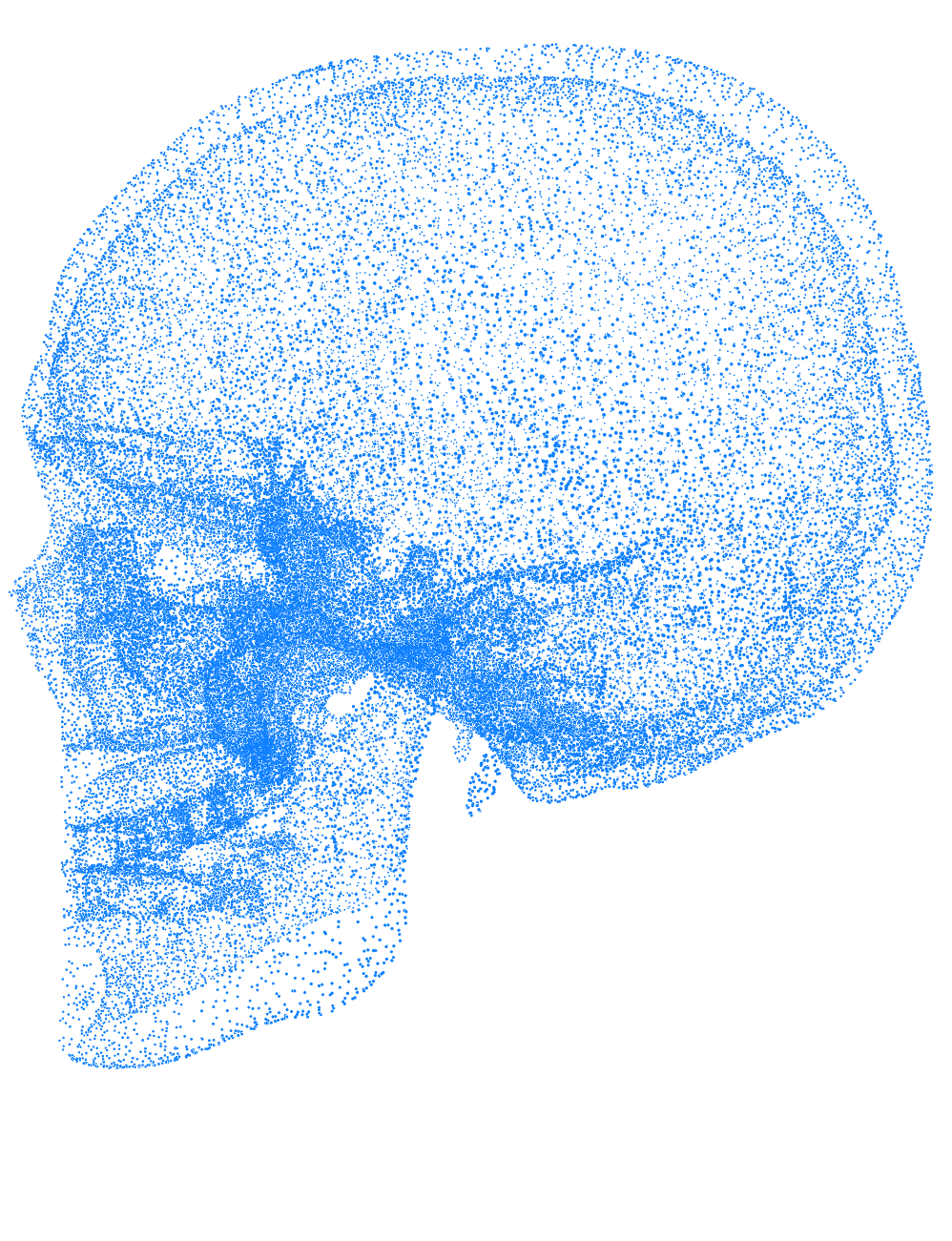}       &
                  \includegraphics[height=55mm]{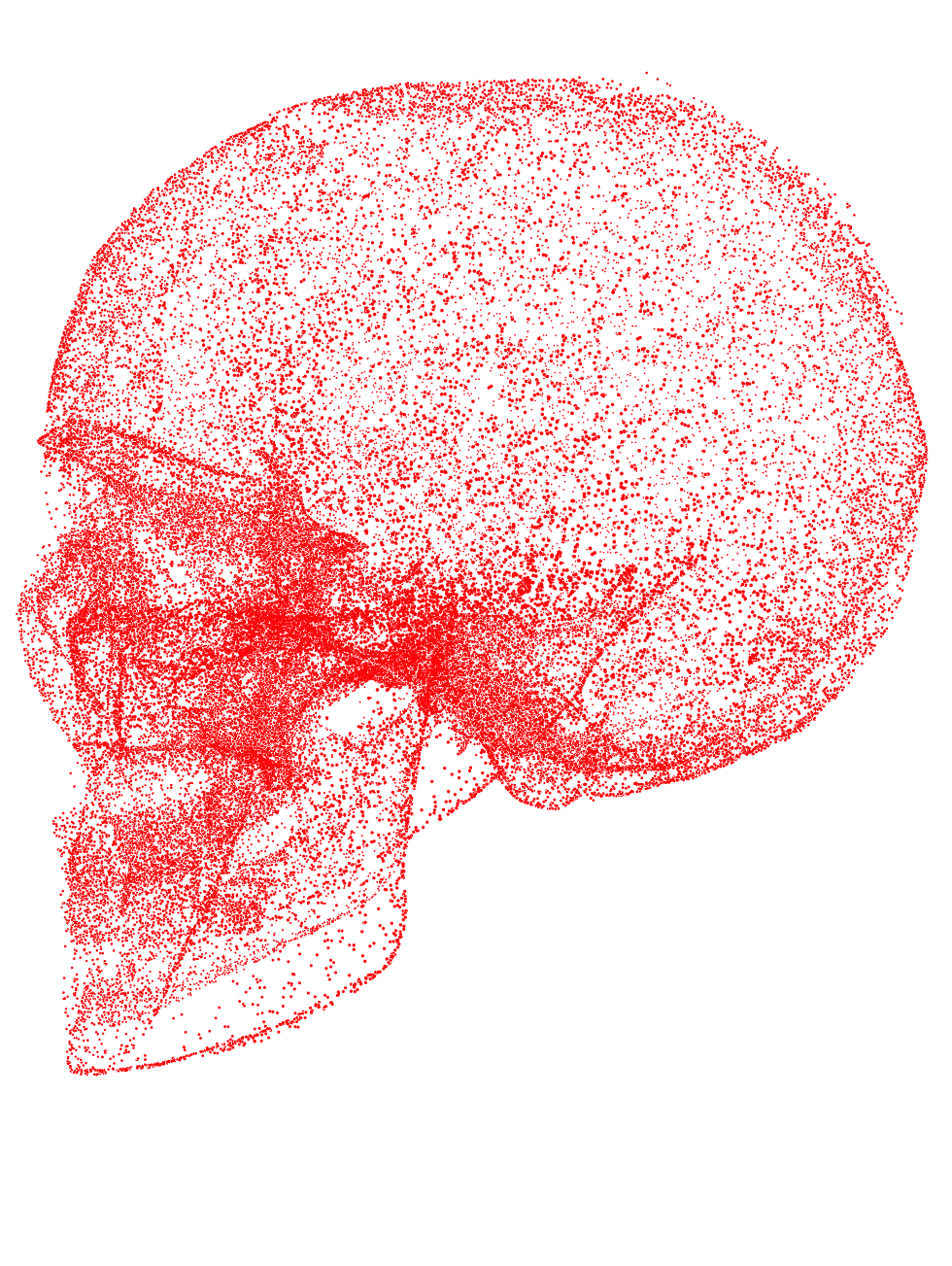}\\\bottomrule
\end{tabular}
\caption{Front and side views on NeuralQAAD reconstructions of Skulls \cite{AchenbachBGHSSB18}. Each skull is compressed to a latent vector of size 10. Low as well as high frequencies can be mostly recovered by NeuralQAAD in contrast to AtlasNetV2 that fails on low frequencies like the skullcap or on high frequencies like the teeth.}
\label{fig:experiments_skulls}
\end{figure*}
We process Skulls in a batch size of 16 instances and use 16 patches for both NeuralQAAD and AtlasNetV2. The visual results can be seen in Figure \ref{fig:experiments_skulls}. NeuralQAAD is able to recover almost all low as well as high frequencies. In contrast, AtlasNetV2 fails in both. From the side view, it can be clearly recognized that detailed and non-manifold structures as teeth and the nasal area are lost by AtlasNetV2 to a huge extent but can be fully reconstructed by NeuralQAAD. Further, rather coarse structures like the eye socket (front view) or the skullcap (front and side view) are badly recovered by AtlasNetV2 whereas NeuralQAAD does not suffer from this. We attribute the better volumetric reconstruction capabilities mostly to the improved training procedure. However, also the effects of weight sharing may be observed. AtlasNetV2 clearly struggles in gluing together individual patches. This does not seem to be an issue for NeuralQAAD that works on shared low level features. The visual impressions are strongly supported by the measured EM-kD losses stated in Table \ref{tab:experiments_skulls}. Solely training on the augmented Chamfer loss already results in an improvement of 21\% in terms of EM-kD. After training with our approach we even note an improvement of 78\%.

\begin{table}[H]
\begin{tabular}{l|ccc}
                  & AtlasNetV2 & \multicolumn{2}{c}{NeuralQAAD} \\ \cline{1-4} 
Trained on        & Aug. Chamfer    & Aug. Chamfer          & QAP         \\ 
EM-kD             & 30.681     & 24.004           & 6.708      
\end{tabular}
\caption{Reconstruction EM-kD for the Skulls dataset. NeuralQAAD performances significantly better than the previous state-of-the-art AtlasNetV2 even if trained on augmented Chamfer loss. Nonetheless, the results suggest that training only on the augmented Chamfer loss is not sufficient.}
\label{tab:experiments_skulls}
\vspace{-7mm}
\end{table}

\begin{figure*}[t]
\centering
\begin{tabular}{ >{\centering\arraybackslash}m{0.14\linewidth} >{\centering\arraybackslash}m{0.14\linewidth} >{\centering\arraybackslash}m{0.14\linewidth} 
                                                      >{\centering\arraybackslash}m{0.14\linewidth} >{\centering\arraybackslash}m{0.14\linewidth} >{\centering\arraybackslash}m{0.14\linewidth} }
\toprule
                \footnotesize NeuralQAAD & \footnotesize Original & \footnotesize AtlasNetV2 & \footnotesize NeuralQAAD & \footnotesize Original & \footnotesize AtlasNetV2 \\\midrule
\footnotesize \includegraphics[height=52mm]{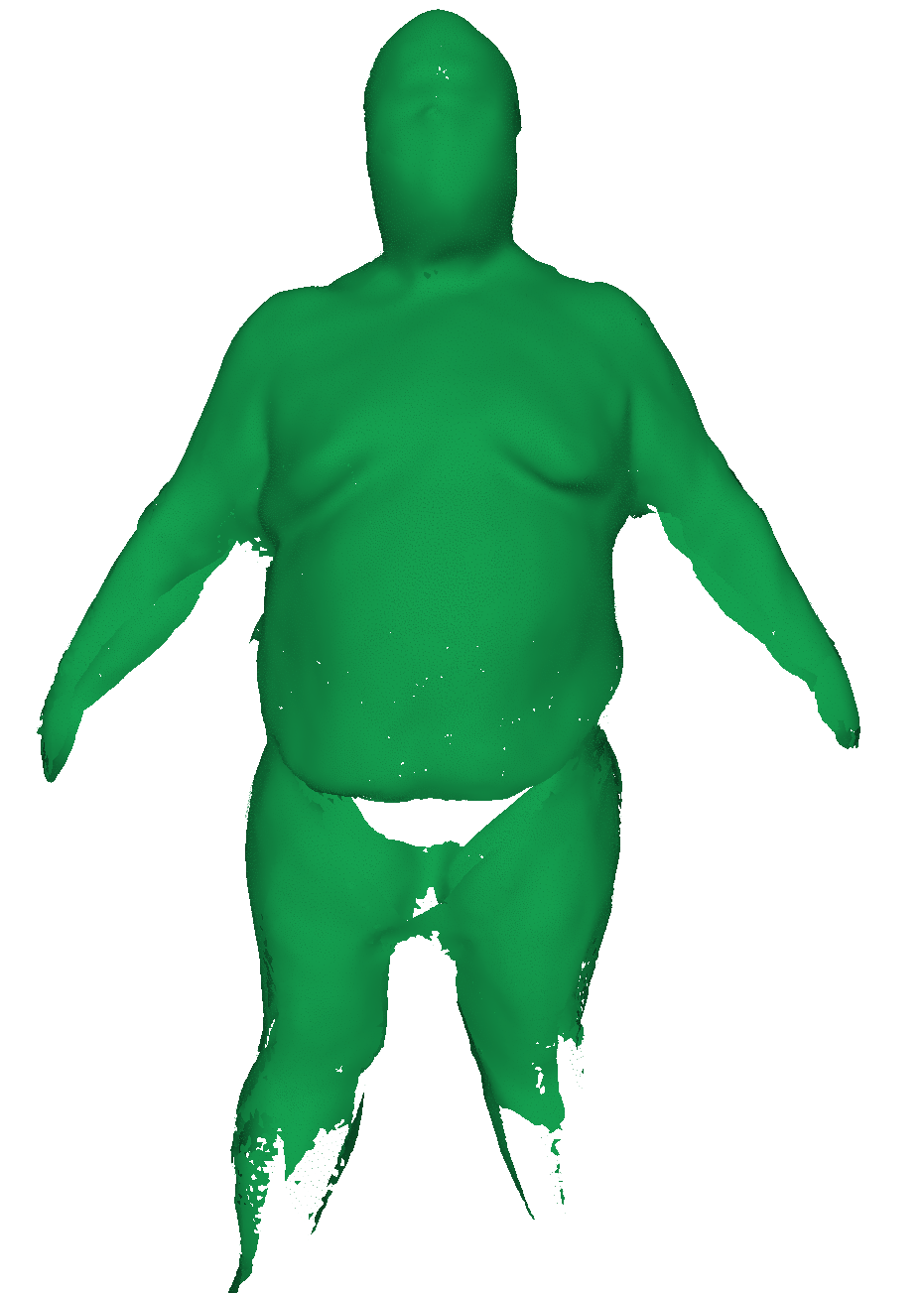}  &
                  \includegraphics[height=52mm]{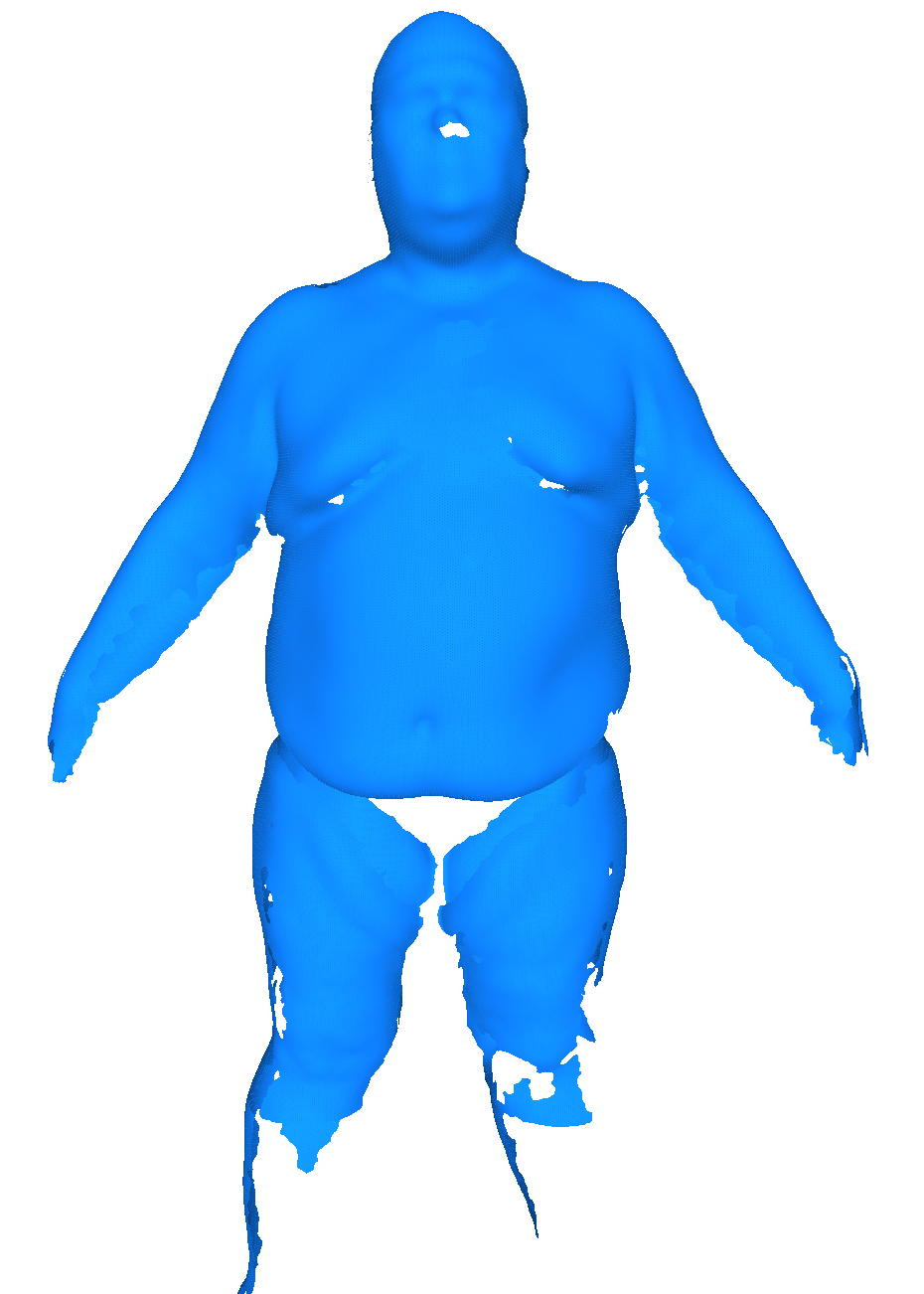}   &
                  \includegraphics[height=52mm]{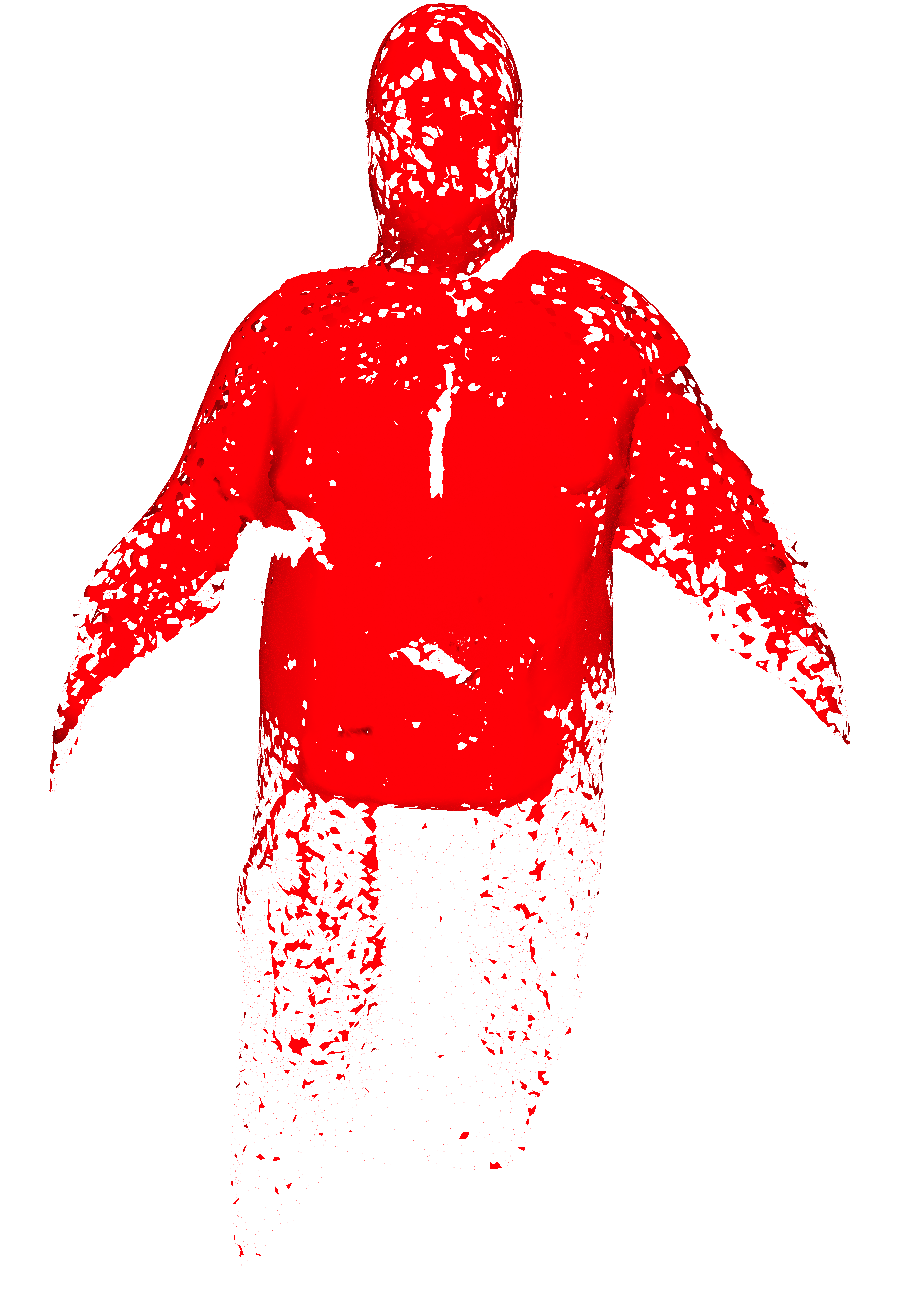}
                & \includegraphics[height=52mm]{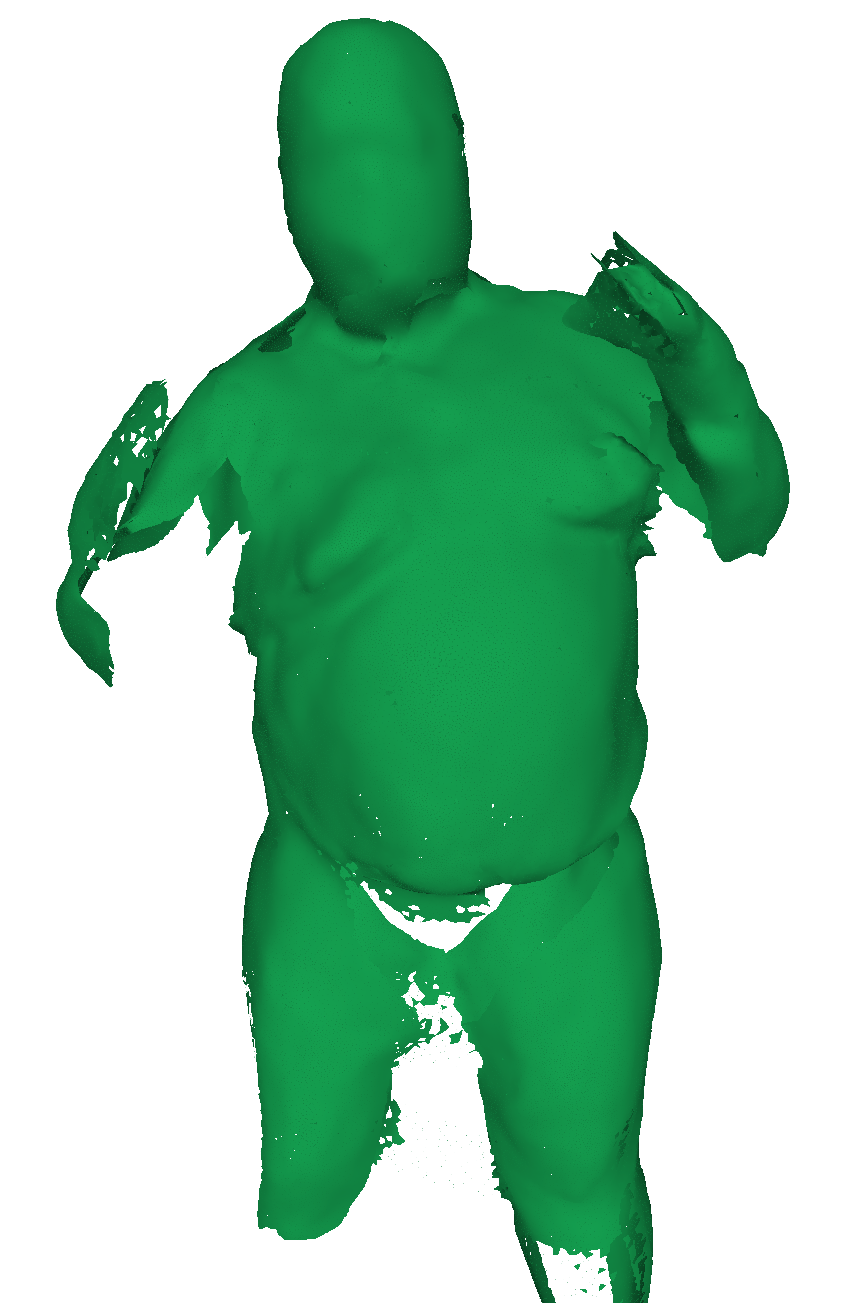}  &
                  \includegraphics[height=52mm]{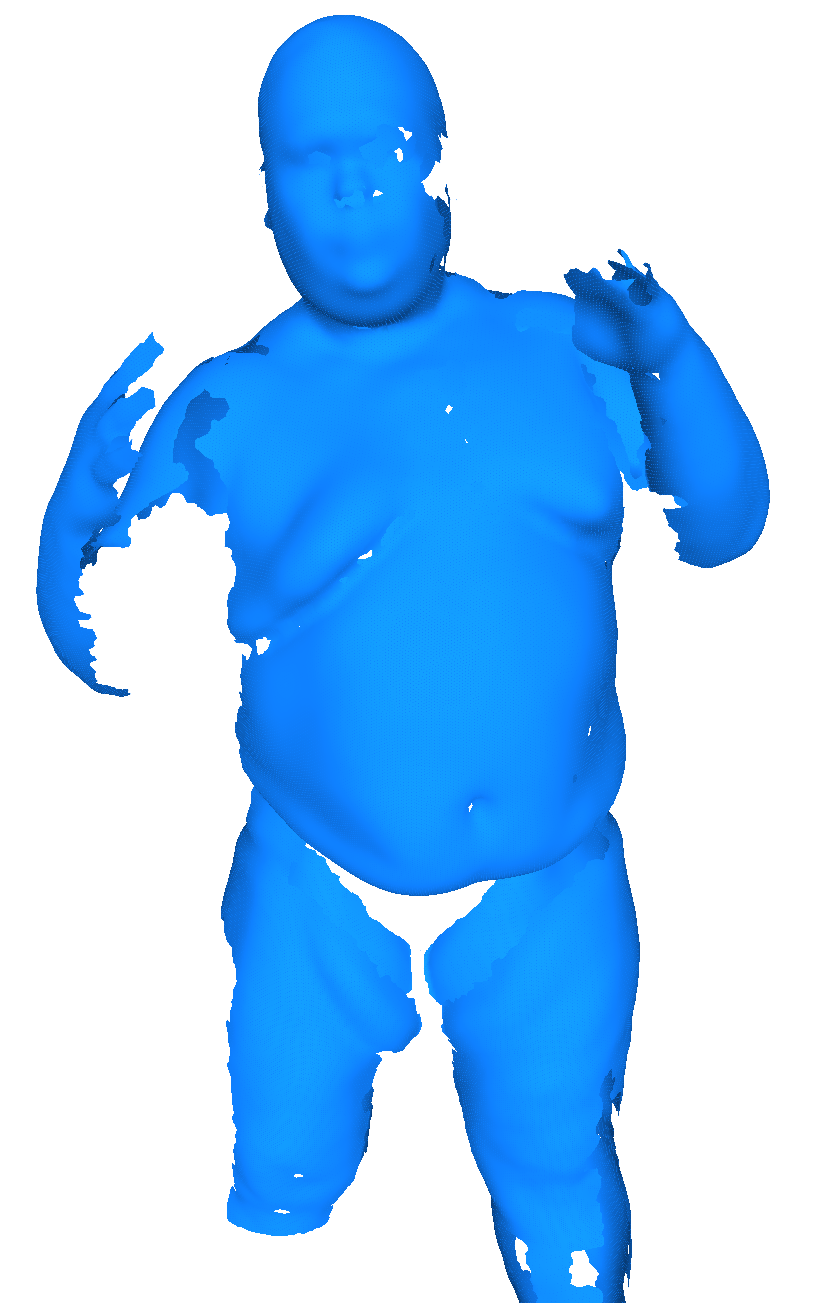}   &
                  \includegraphics[height=52mm]{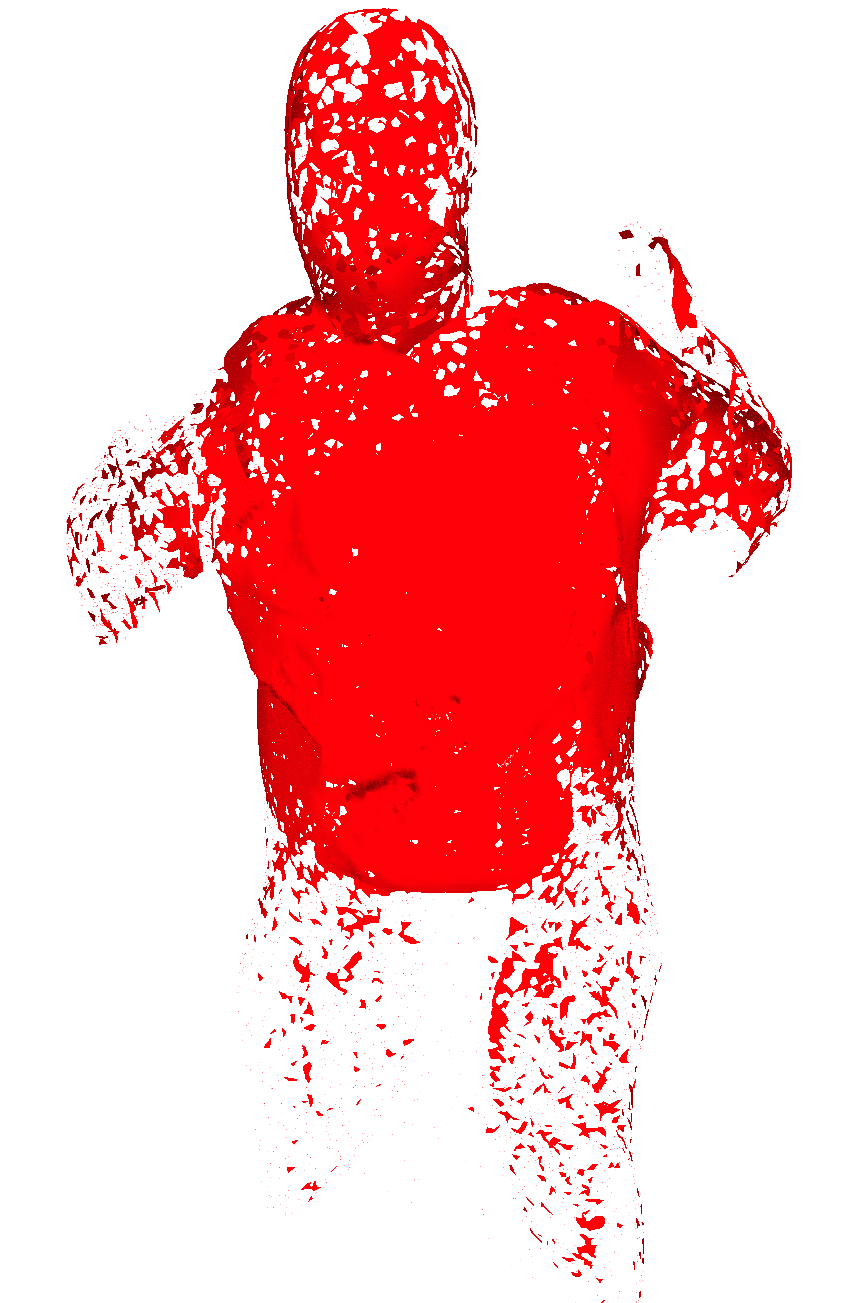}\\

\footnotesize \includegraphics[height=52mm]{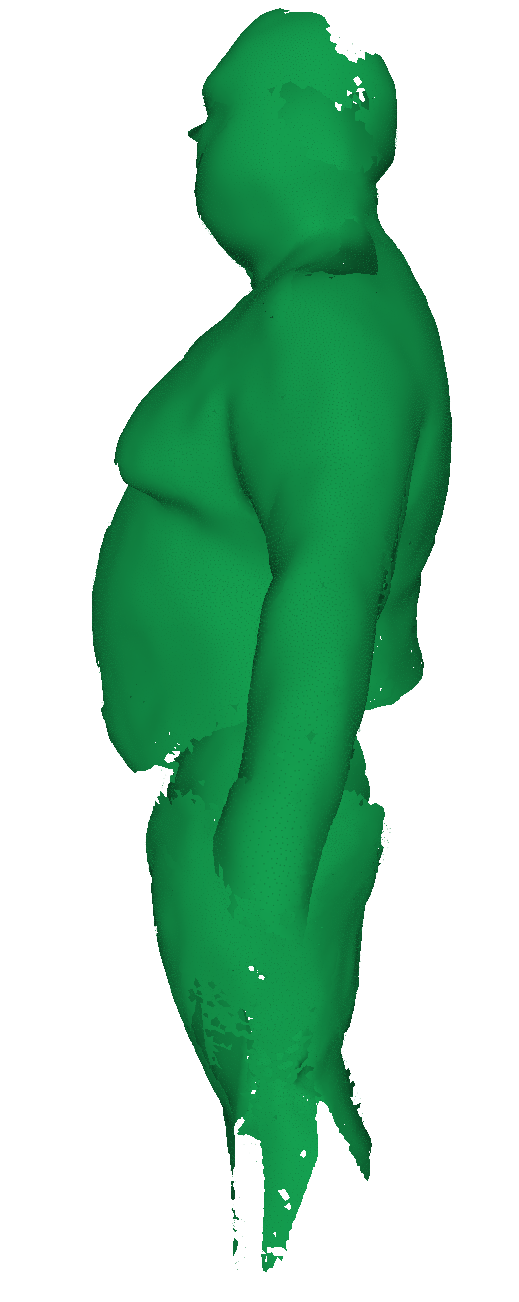}  &
                  \includegraphics[height=52mm]{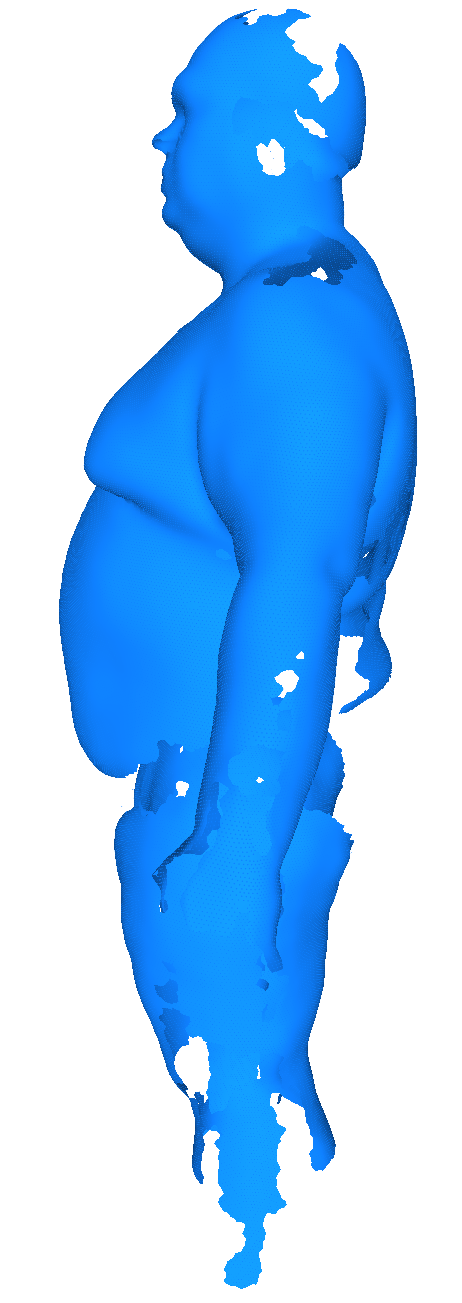}   &
                  \includegraphics[height=52mm]{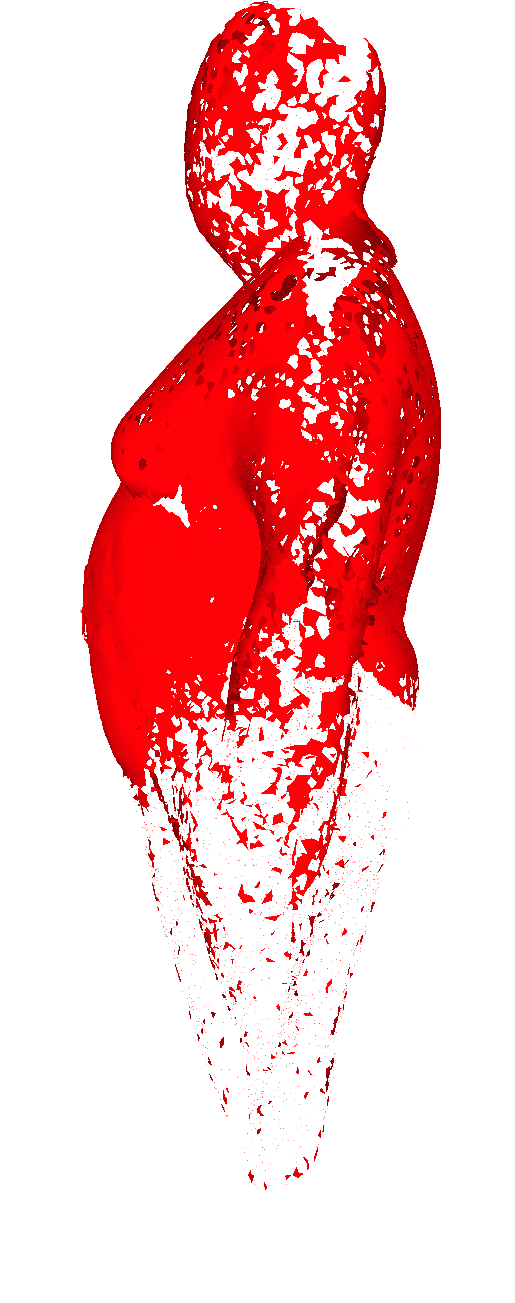}
                & \includegraphics[height=52mm]{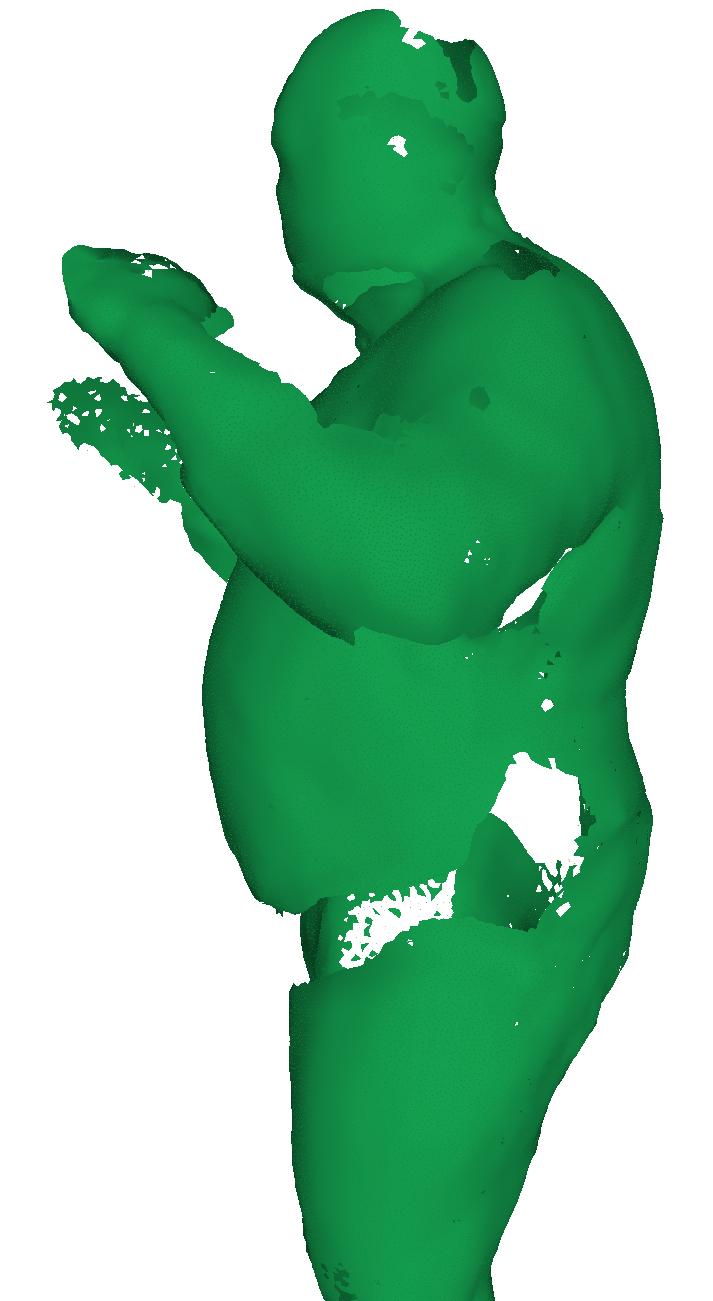}  &
                  \includegraphics[height=52mm]{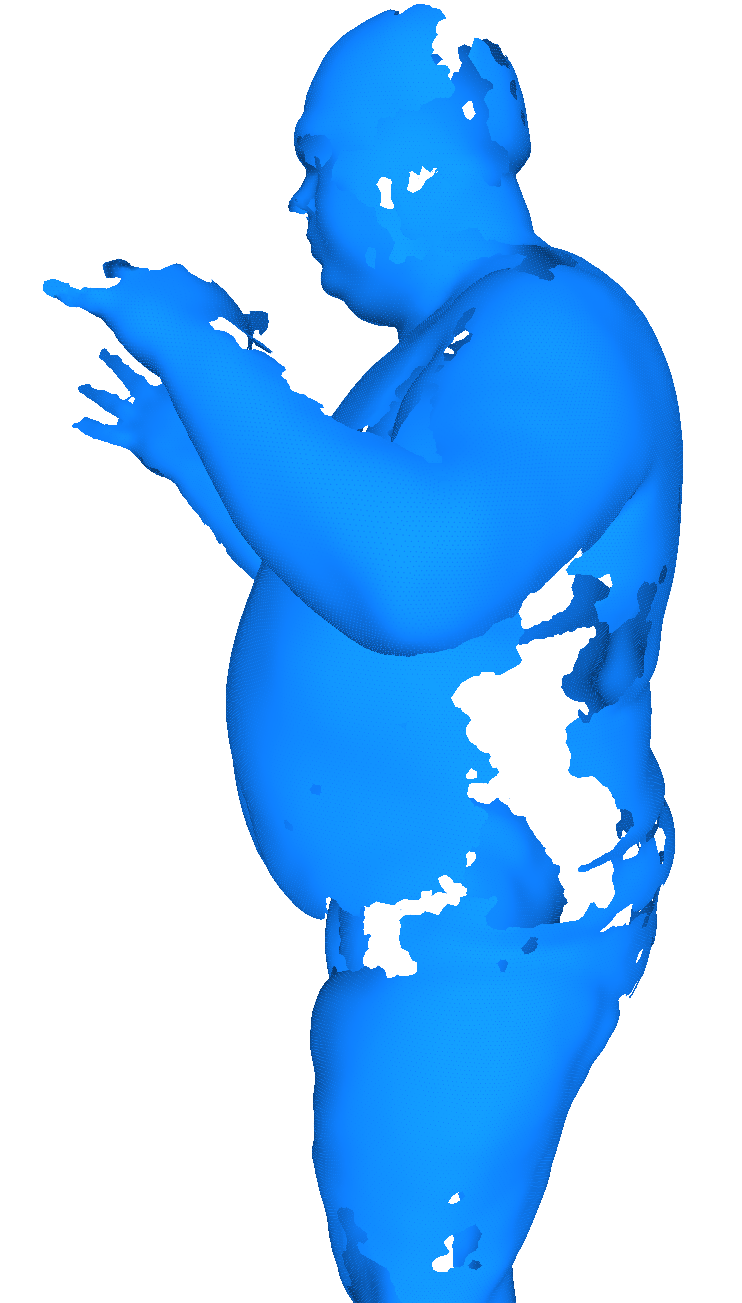}   &
                  \includegraphics[height=52mm]{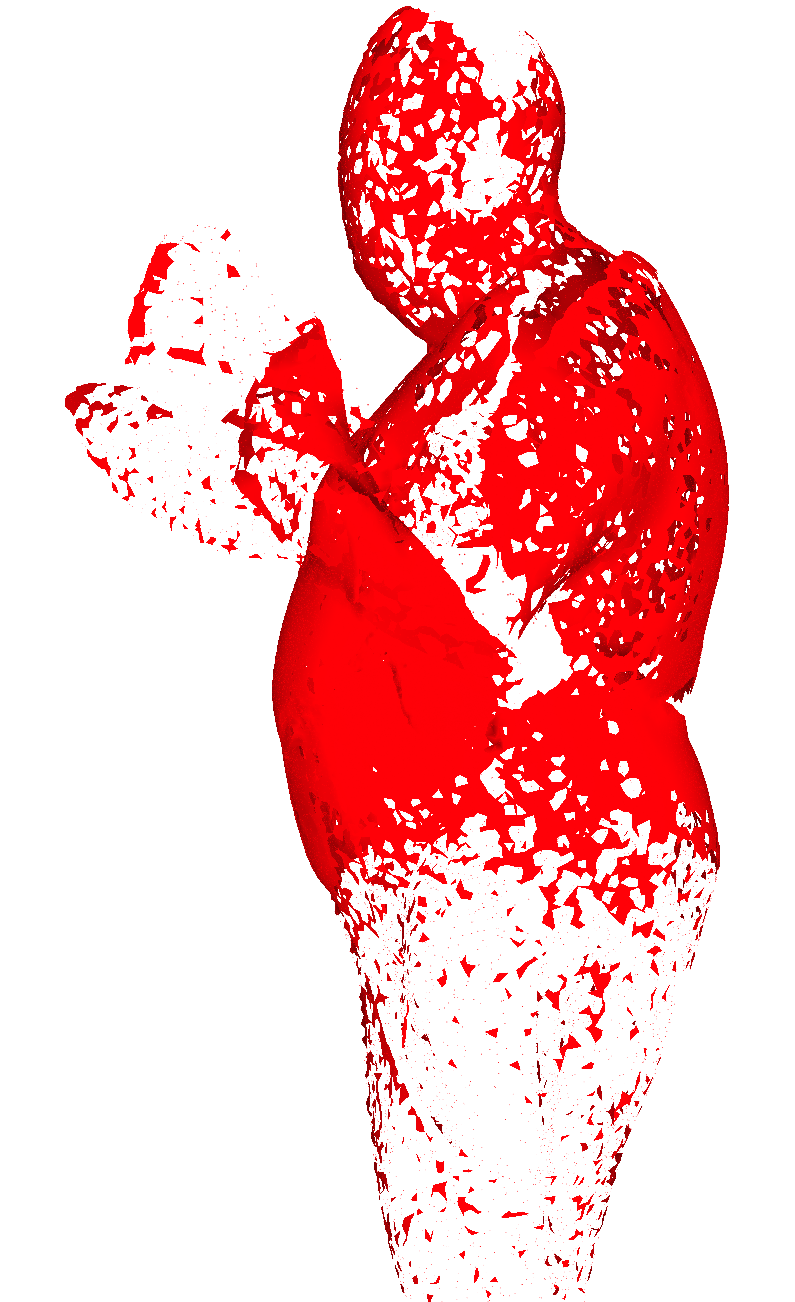}\\\bottomrule
\end{tabular}
\caption{Front and side views on NeuralQAAD reconstructions of D-Faust \cite{Dfaust}. Each body is compressed to a latent vector of size 256. The differences between NeuralQAAD and AtlasNetV2 are clearly visible as AtlasNetV2 is not able to recover even coarse structures like extremities.}
\label{fig:experiments_dfaust}
\end{figure*}

\textbf{D-Faust:} The D-Faust dataset is mostly described in low frequency changes. We compress each instance to a latent code of size 256. Training is conducted with batches of size 16 while using 128 patches for NeuralQAAD. To ensure comparability AtlasNetV2 can only make use of 28 patches. The visual results can be seen in Figure \ref{fig:experiments_dfaust}. Contrary to the volumetric Skulls dataset, we can restore a surface for each point cloud to ease the visual comparison of reconstruction results. For this, the simple ball-pivoting algorithm \cite{ballpiv} is used with identical hyperparameters for all experiments.

\begin{figure*}[t]
\centering
\begin{tabular}{ >{\centering\arraybackslash}m{0.3\linewidth} >{\centering\arraybackslash}m{0.3\linewidth} >{\centering\arraybackslash}m{0.3\linewidth} }
\toprule
              \footnotesize Skulls & \footnotesize D-Faust & \footnotesize COMA \\\midrule
\footnotesize \includegraphics[height=42mm]{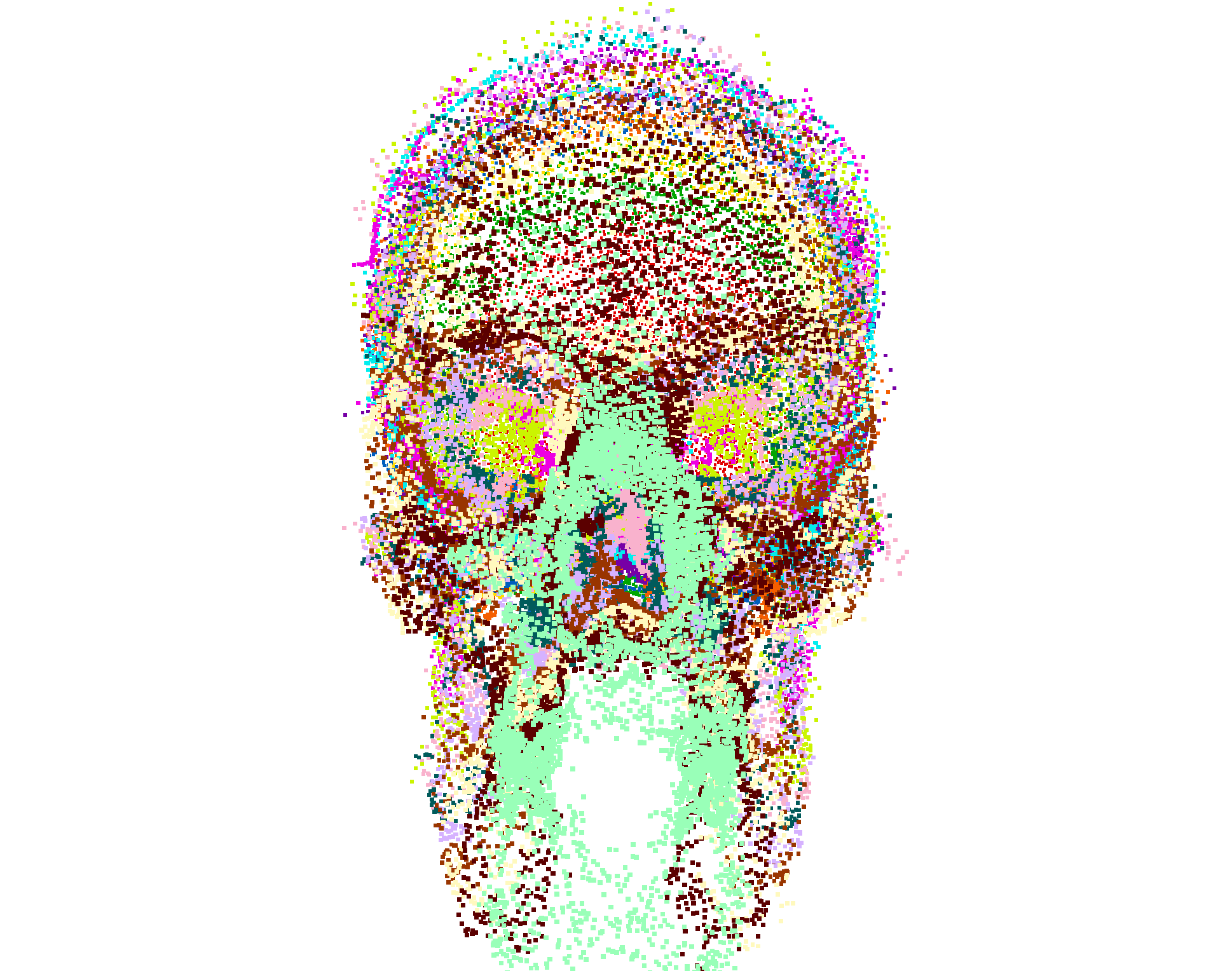}  &
                  \includegraphics[height=42mm]{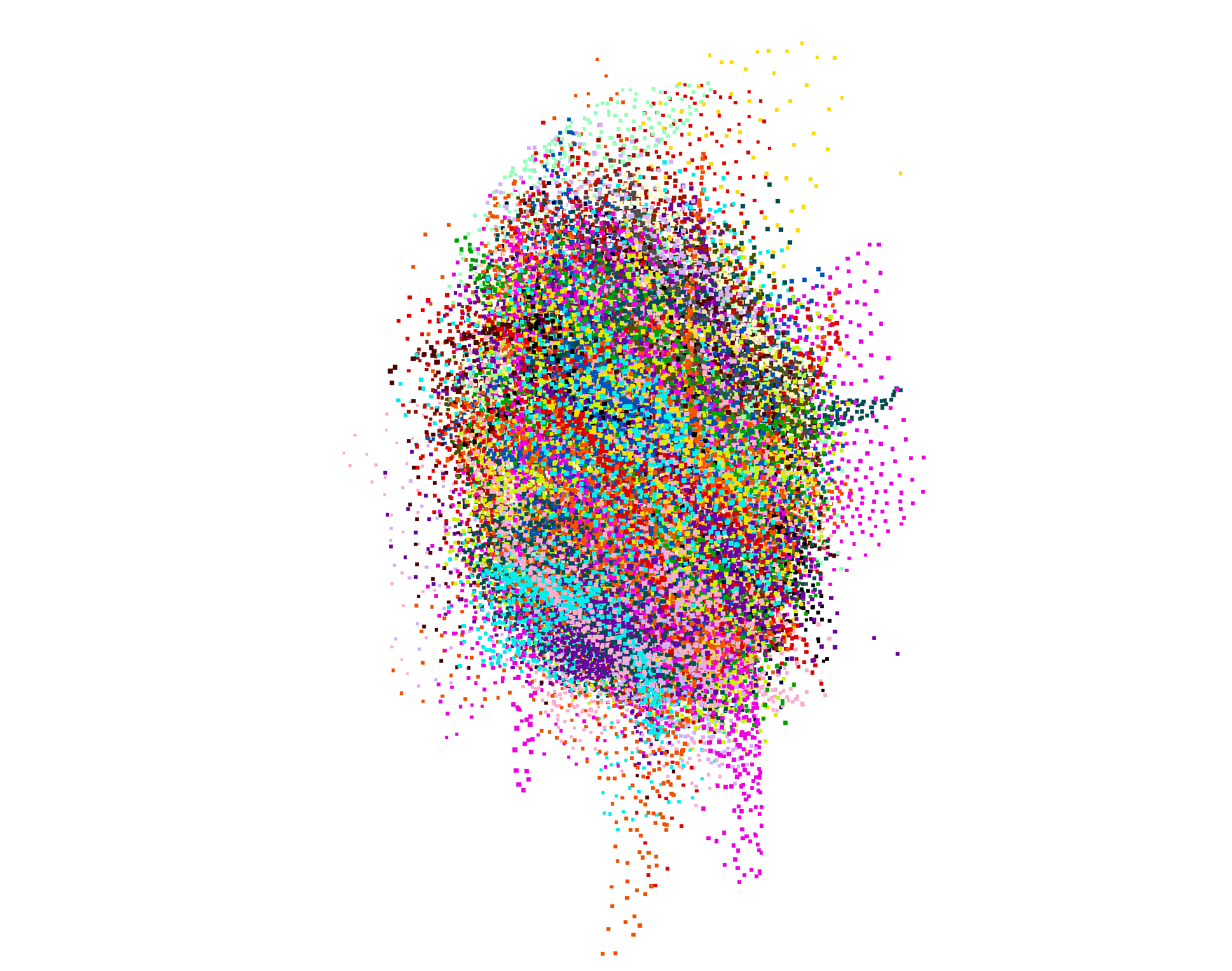}   &
                  \includegraphics[height=42mm]{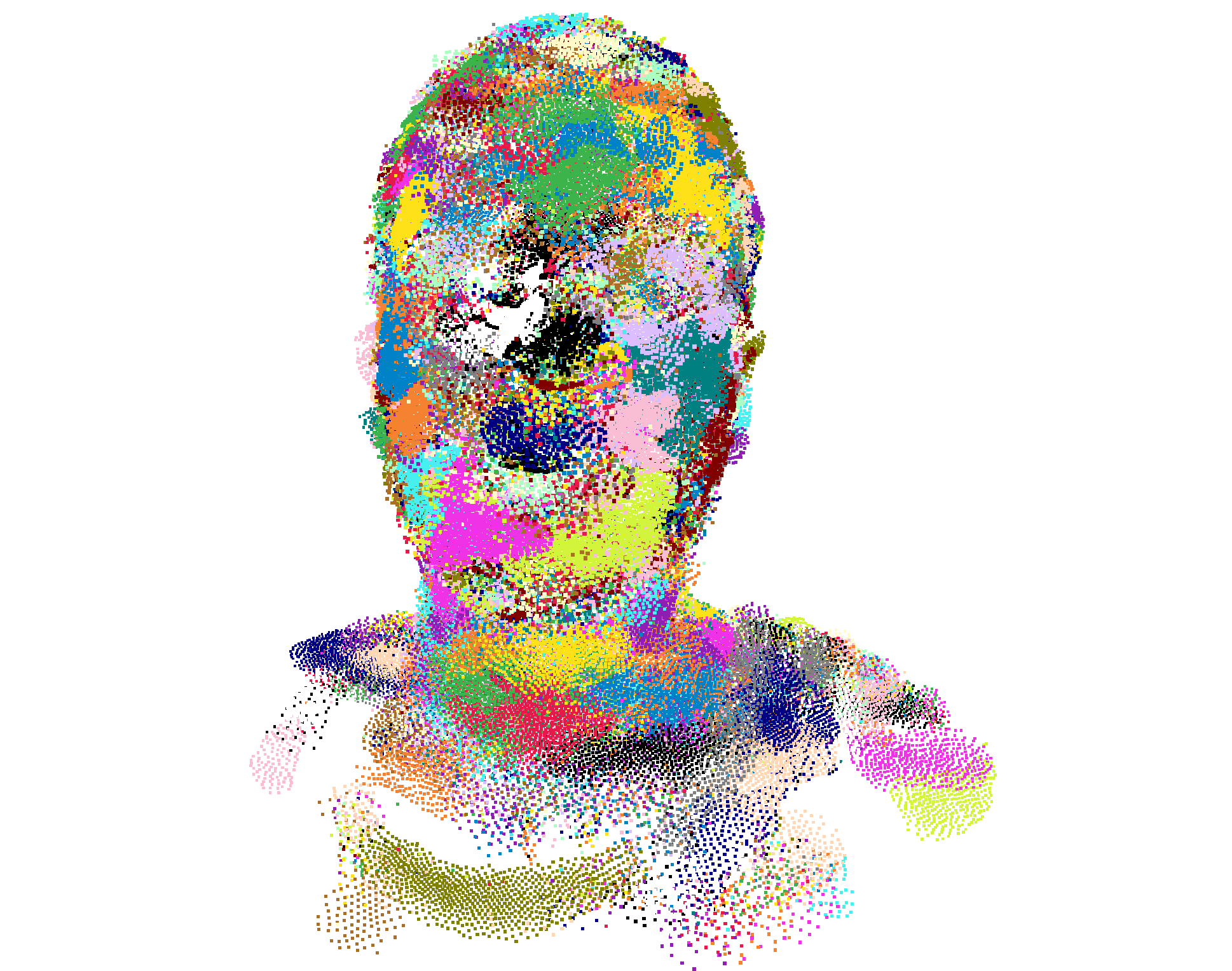}\\\bottomrule
\end{tabular}
\caption{The trainable input point clouds after QAP training for each evaluated dataset. Colors reflect assignment to patches. For Skulls and COMA, datasets mostly defined by changes in high frequencies, the input point clouds create the impression of noisy instances. On the other hand, D-Faust, mostly defined by changes in low frequencies, exhibits a fully deformed input point cloud. }
\label{fig:experiments_inputs}
\end{figure*}

\begin{table}[H]
\begin{tabular}{l|ccc}
                  & AtlasNetV2 & \multicolumn{2}{c}{NeuralQAAD} \\ \cline{1-4} 
Trained on        & Aug. Chamfer    & Aug. Chamfer          & QAP         \\ 
EM-kD             & 0.076     & 0.020           & 0.010     
\end{tabular}
\caption{Reconstruction EM-kD for the D-Faust dataset. As with Skulls, NeuralQAAD achieves significant improvements with as well as without QAP training. However, the results indicate that the scalability of NeuralQAAD in the number of patches is the main performance factor.}
\label{tab:experiments_dfaust}
\vspace{-7mm}
\end{table}

From all perspectives, it becomes evident that AtlasNetV2 merges even coarse structured extremities and, as with Skulls, struggles to stitch together individual patches. NeuralQAAD, by contrast, is able to almost fully recover all prevalent structures and forming a smooth surface across patch borders. The EM-kD reconstruction losses stated in Tab.~\ref{tab:experiments_dfaust} show a significant improvement through NeuralQAAD. However, they tell a different story as before. Already after training NeuralQAAD with the augmented Chamfer loss, the EM-kD decreases by about 73\% percent. Further, our training scheme leads to a total drop in EM-kD of 86\%. Combined, both numbers indicate that the scalability of NeuralQAAD in the number of patches is the dominant factor of performance for D-Faust whereas the training scheme plays a minor role. The D-Faust experiments reveal another interesting observation. In Figure~\ref{fig:experiments_inputs} the trainable input point clouds are shown. Although initialized with a random example of the respective dataset, only for D-Faust a full deformation can be noticed. The \emph{egg-like} structure seems to be favorable if strong low frequency changes occur which is not the case for Skulls and COMA (see next Section).

\begin{figure*}[t]
\centering
\begin{tabular}{ >{\centering\arraybackslash}m{0.14\linewidth} >{\centering\arraybackslash}m{0.14\linewidth} >{\centering\arraybackslash}m{0.14\linewidth} 
                                                      >{\centering\arraybackslash}m{0.14\linewidth} >{\centering\arraybackslash}m{0.14\linewidth} >{\centering\arraybackslash}m{0.14\linewidth} }
\toprule
                \footnotesize NeuralQAAD & \footnotesize Original & \footnotesize AtlasNetV2 & \footnotesize NeuralQAAD & \footnotesize Original & \footnotesize AtlasNetV2 \\\midrule
\footnotesize \includegraphics[height=35mm]{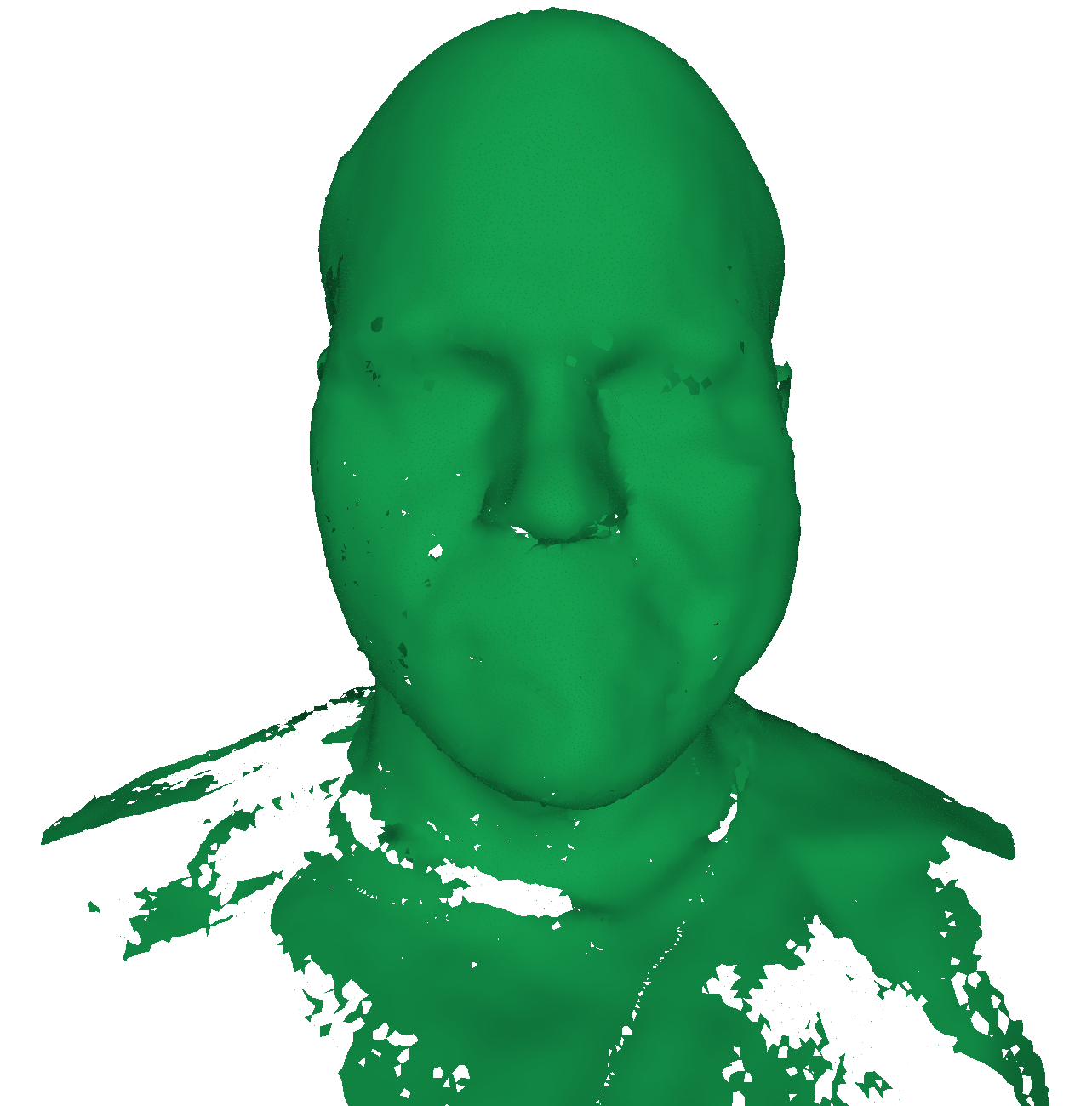}  &
                  \includegraphics[height=35mm]{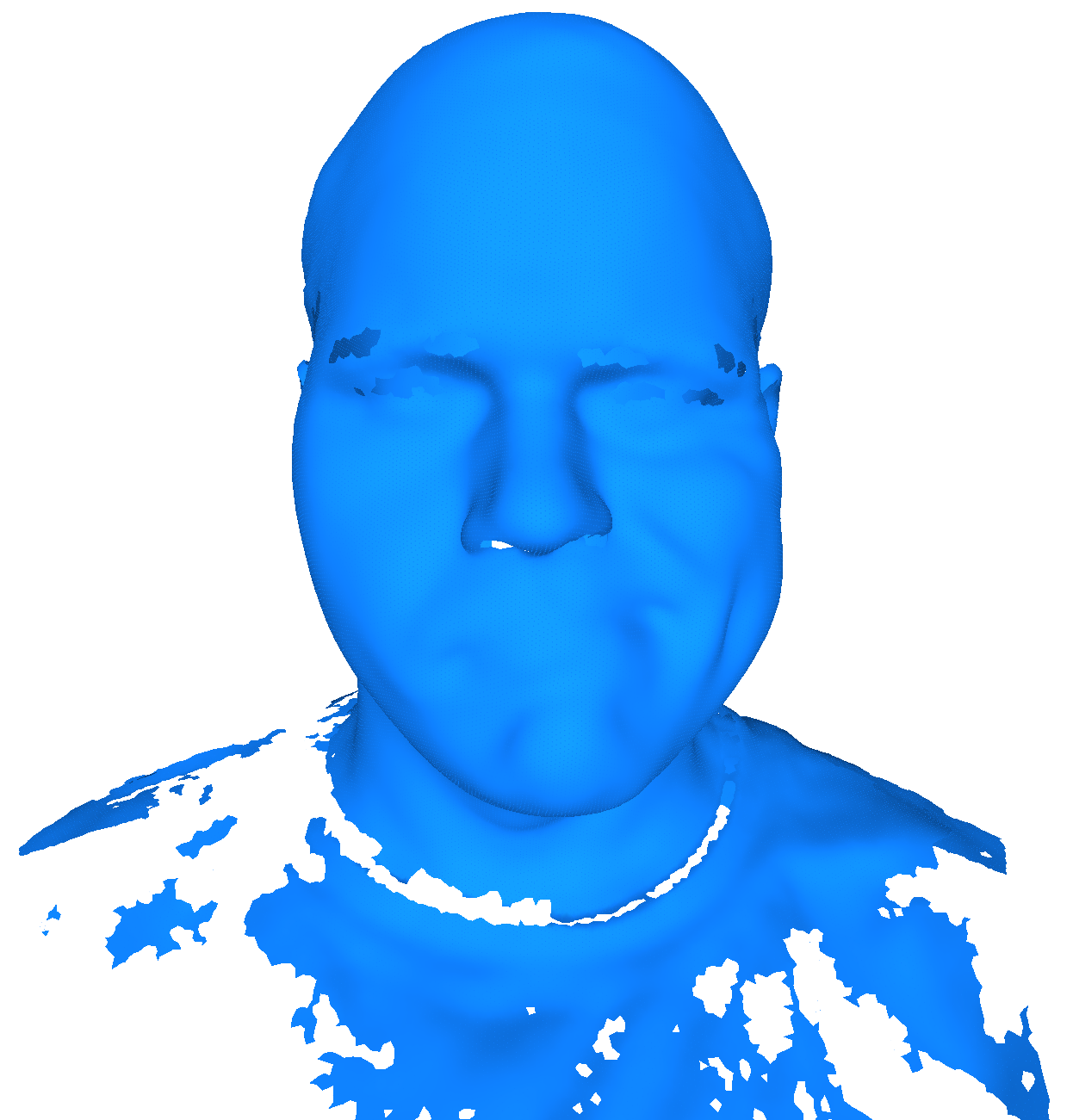}   &
                  \includegraphics[height=35mm]{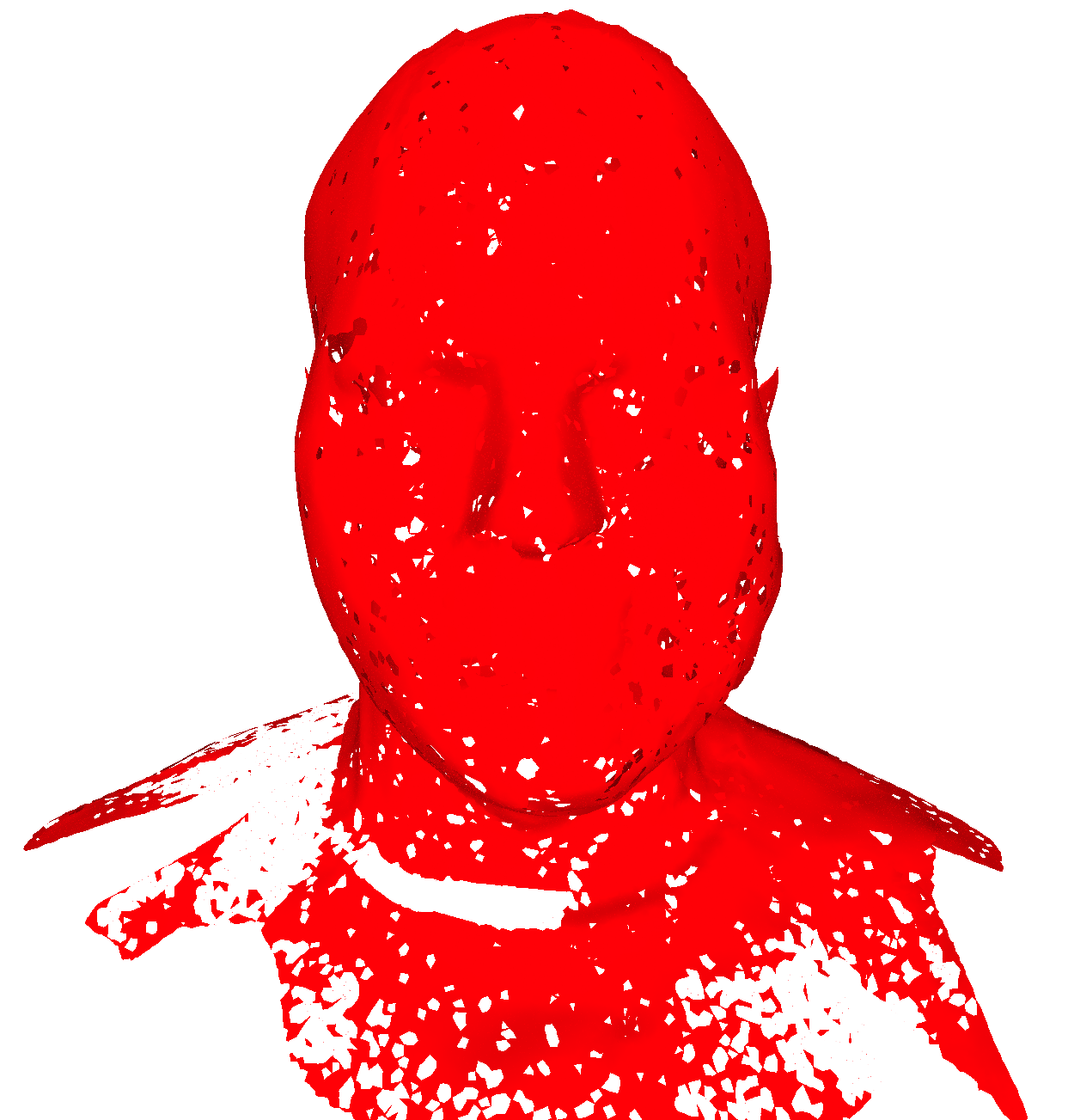}
                & \includegraphics[height=35mm]{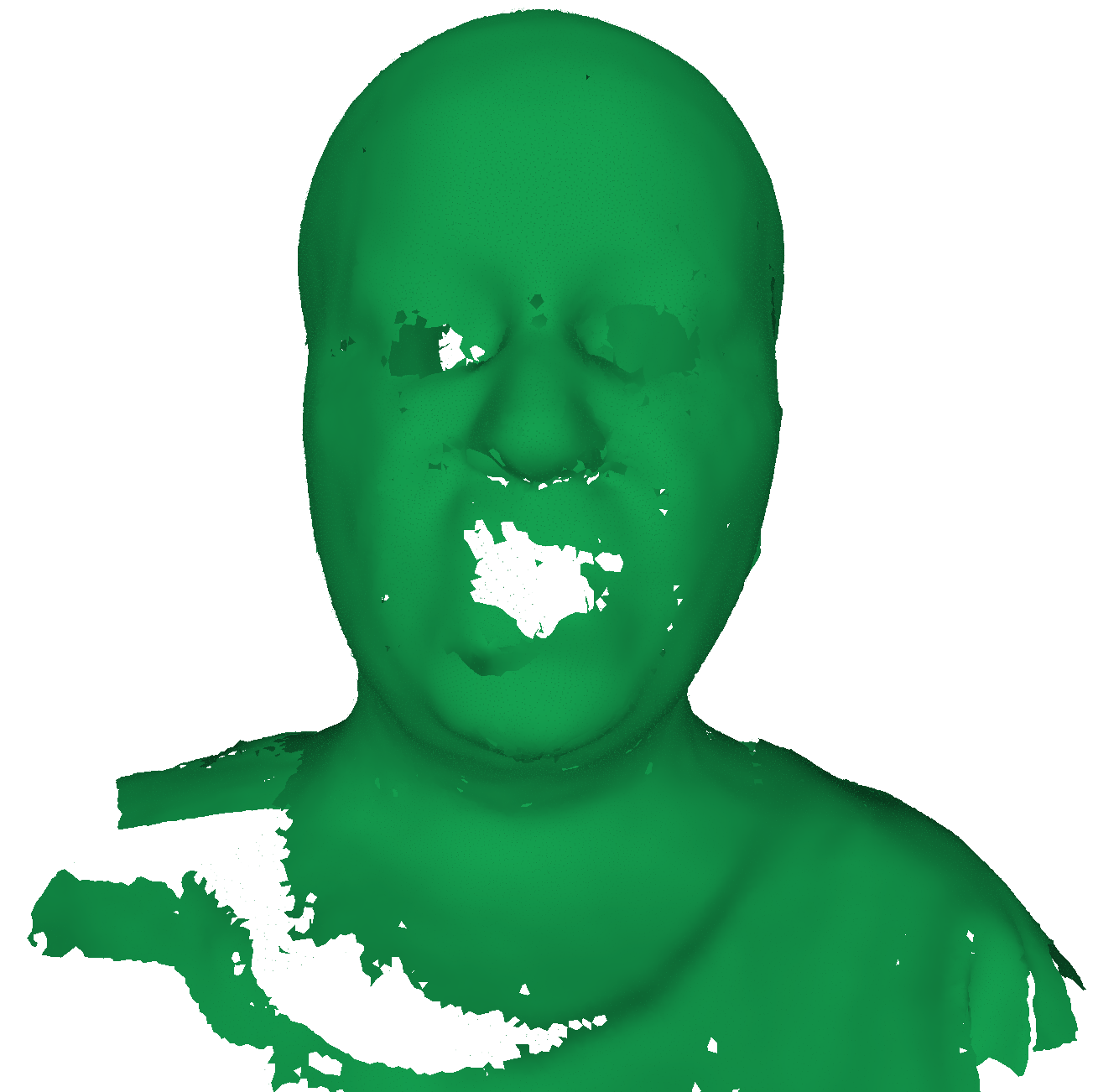}  &
                  \includegraphics[height=35mm]{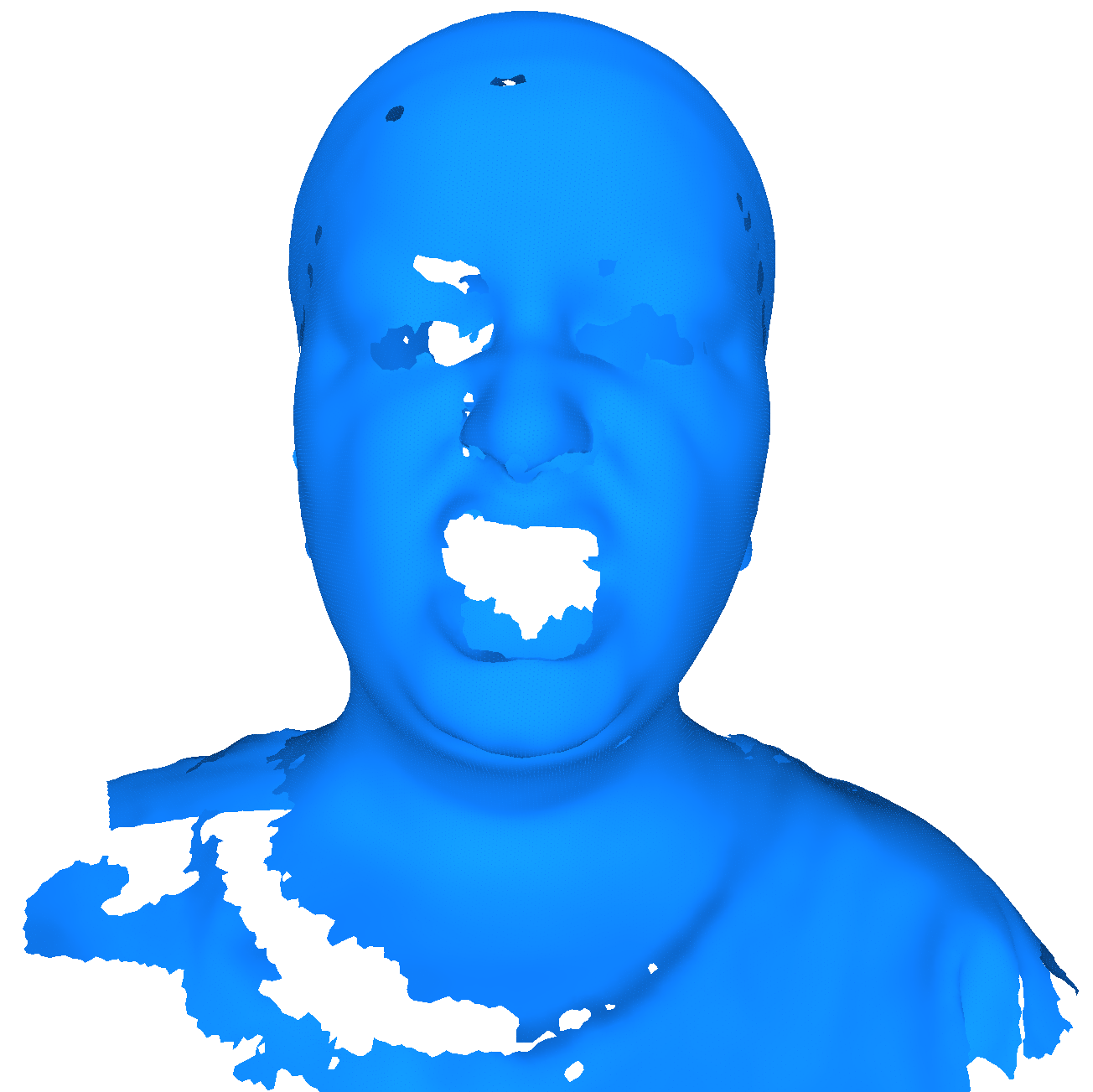}   &
                  \includegraphics[height=35mm]{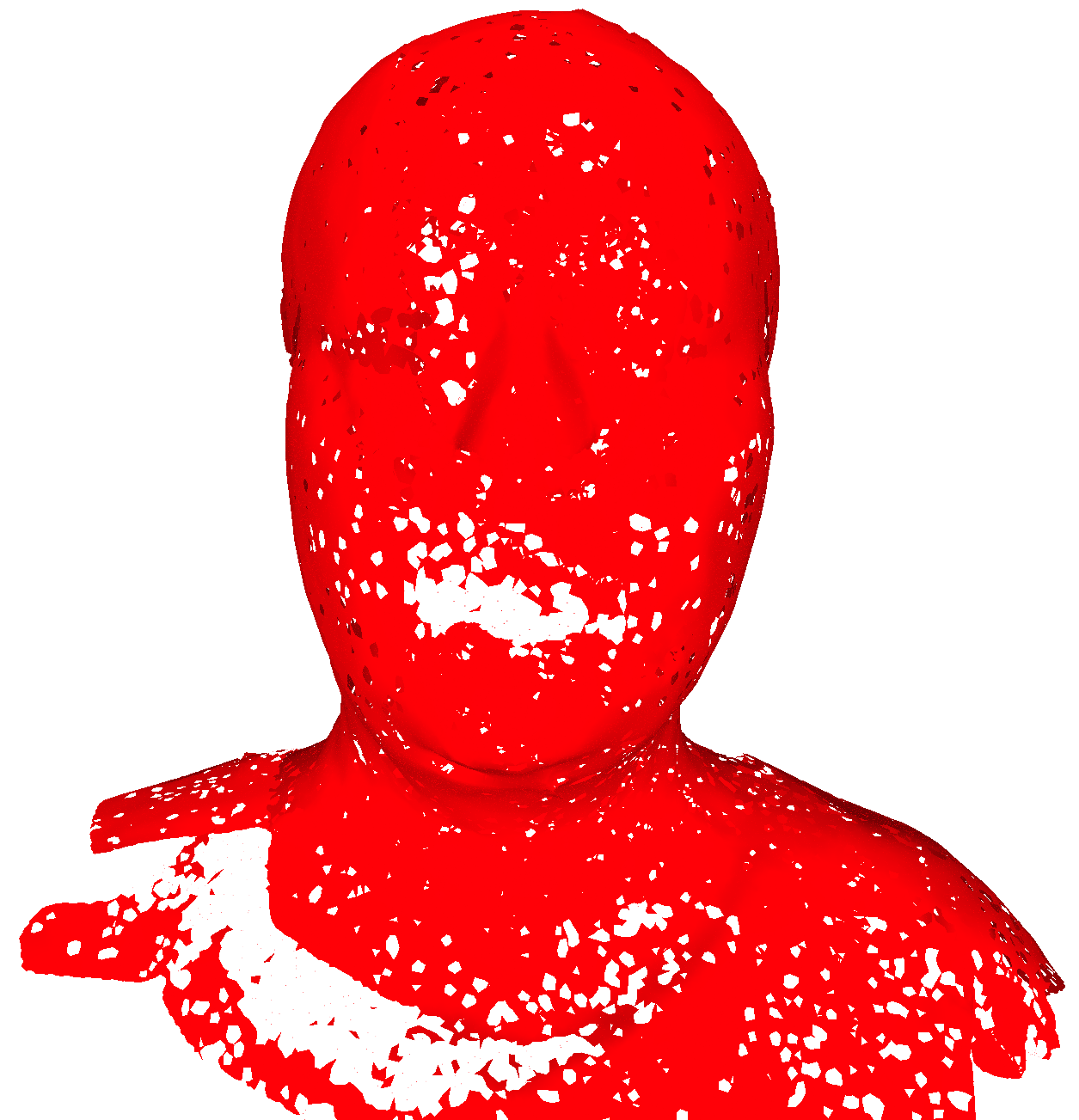}\\

\footnotesize \includegraphics[height=35mm]{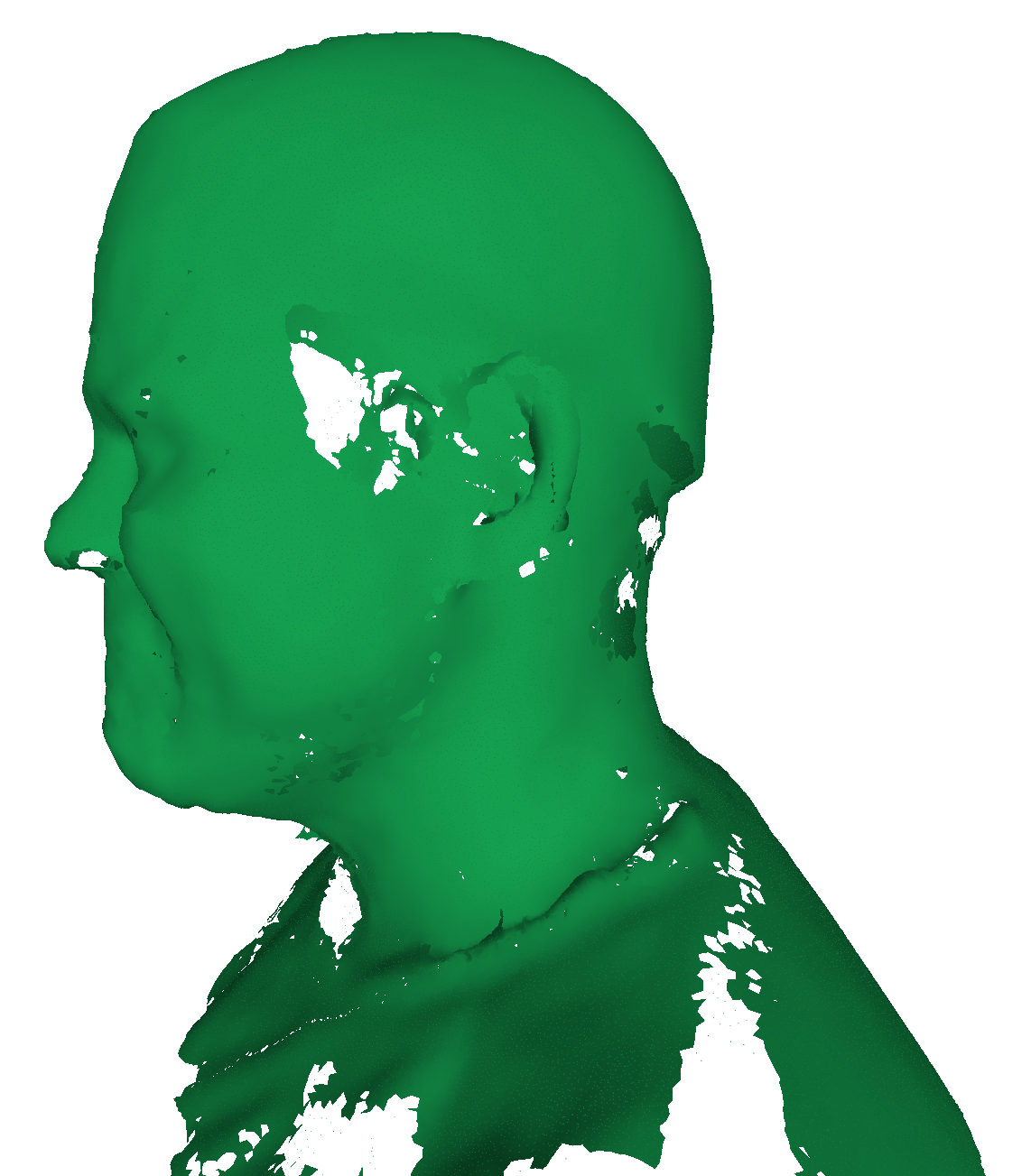}  &
                  \includegraphics[height=35mm]{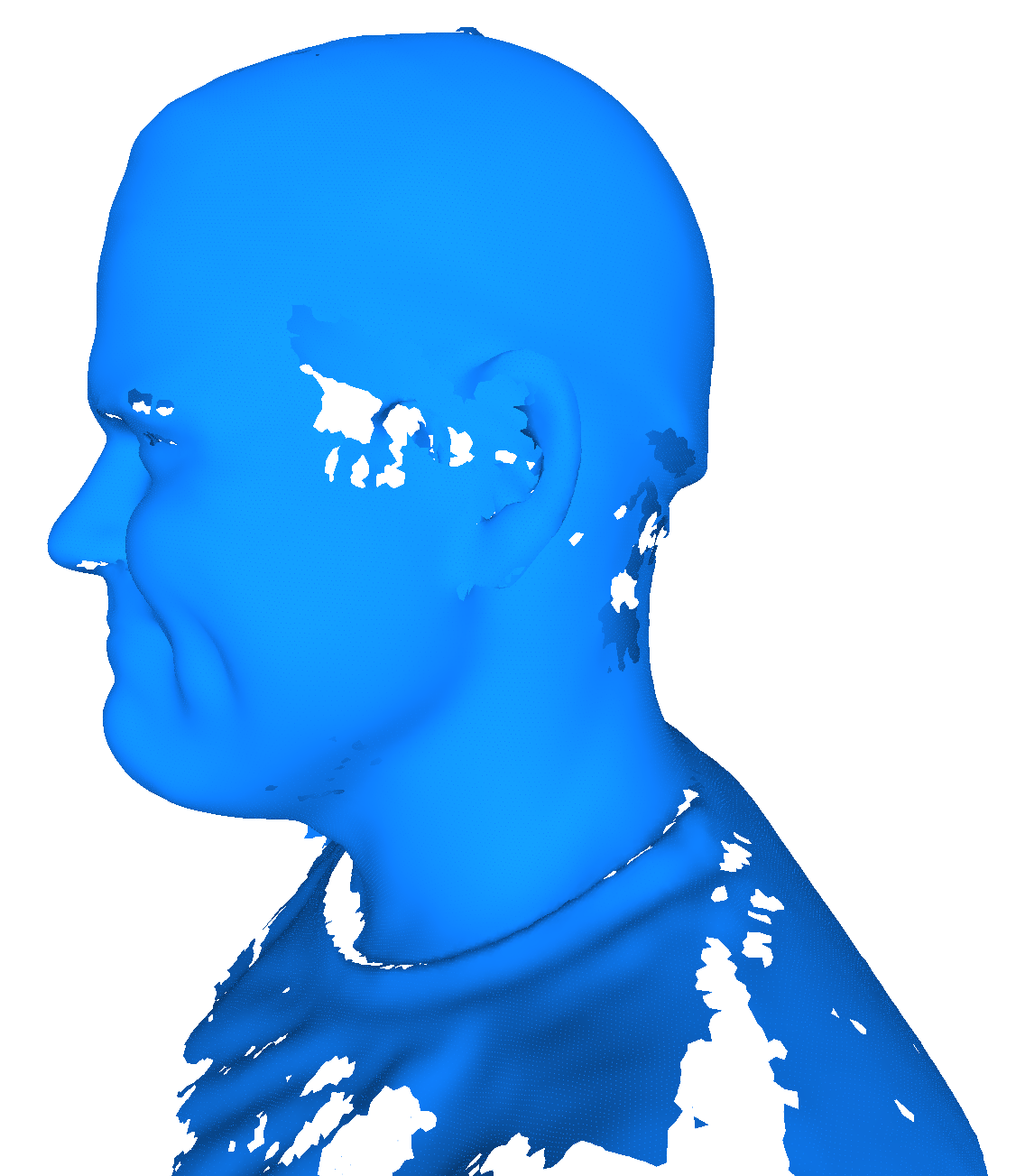}   &
                  \includegraphics[height=35mm]{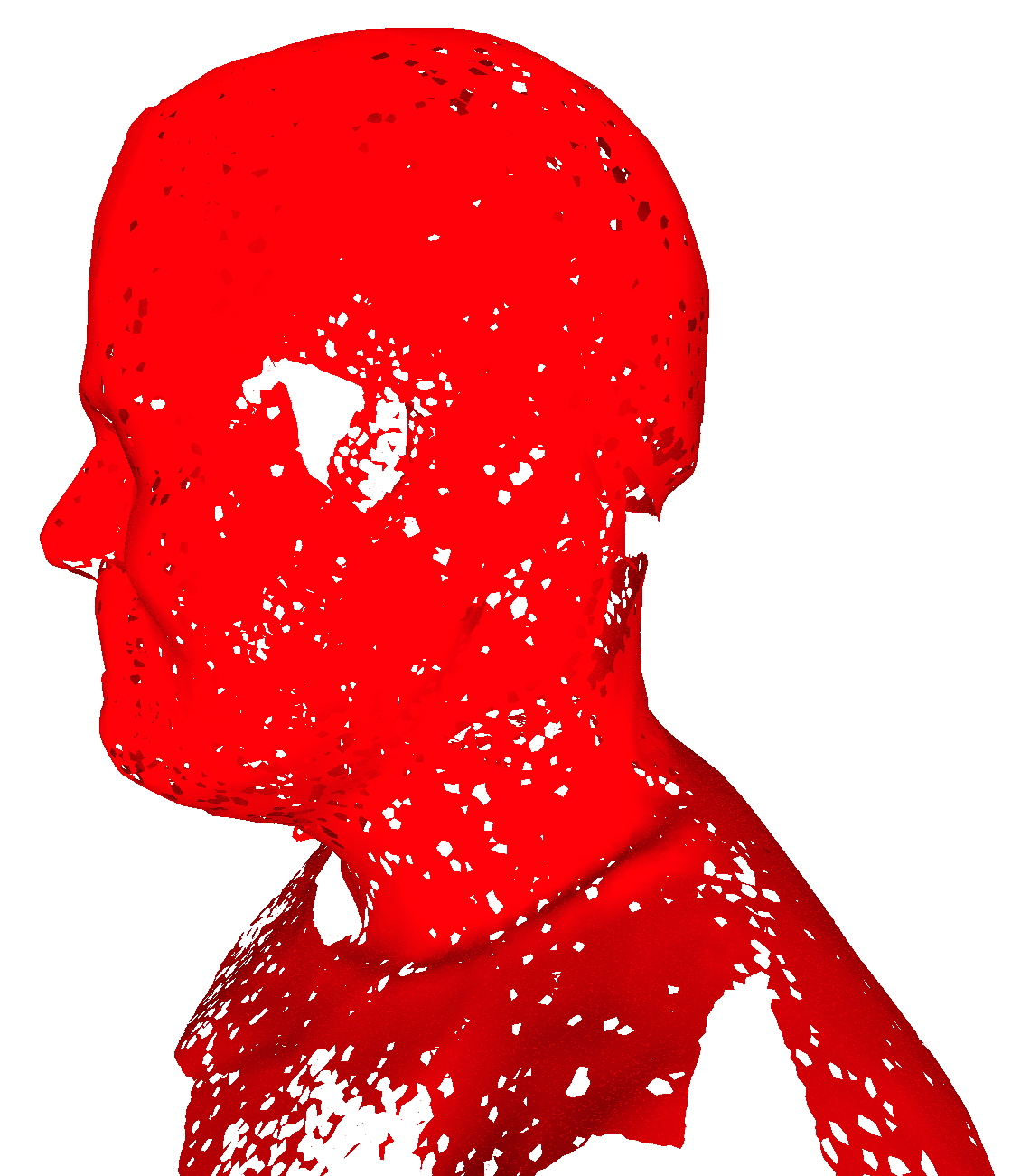}
                & \includegraphics[height=35mm]{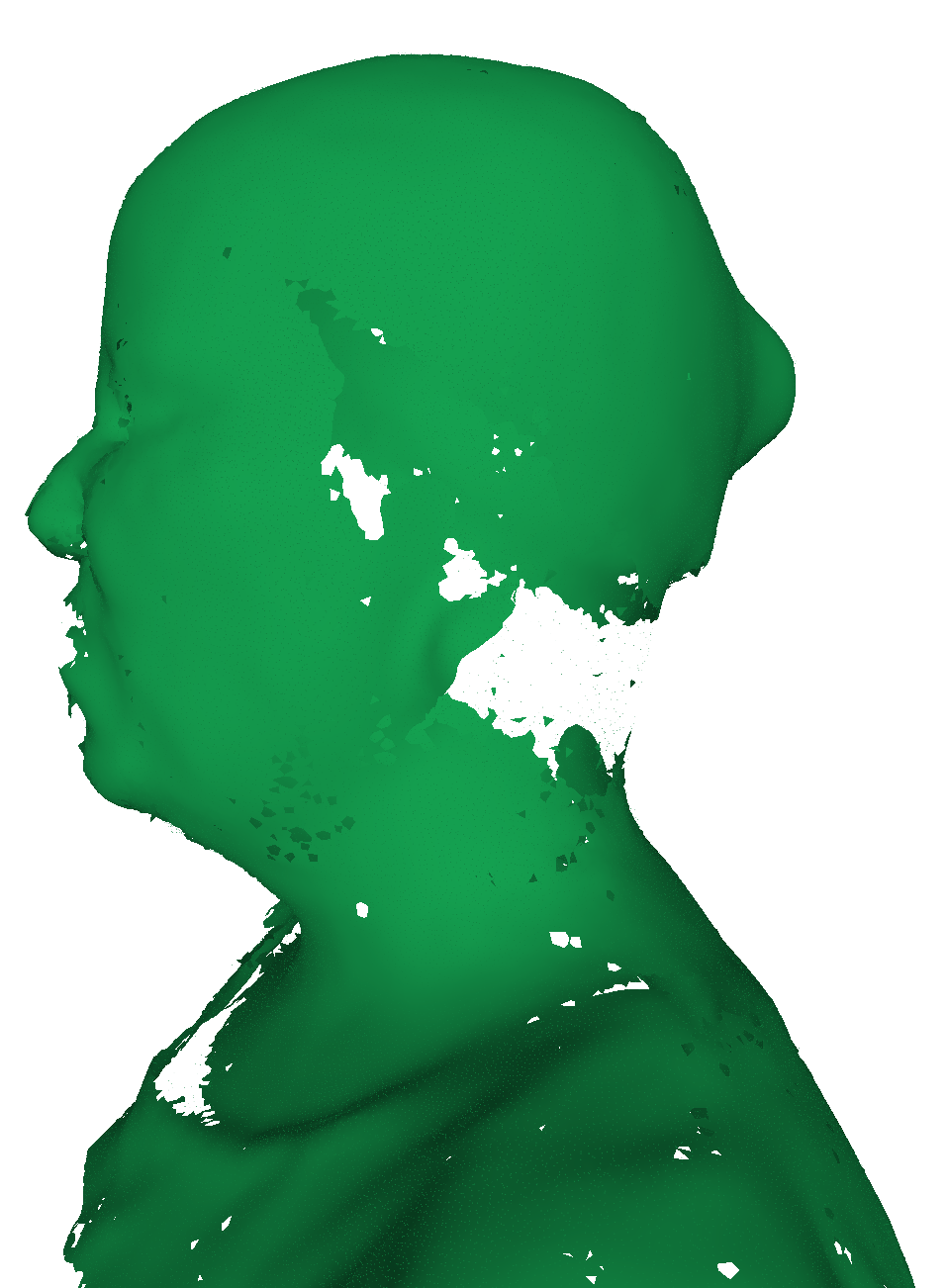}  &
                  \includegraphics[height=35mm]{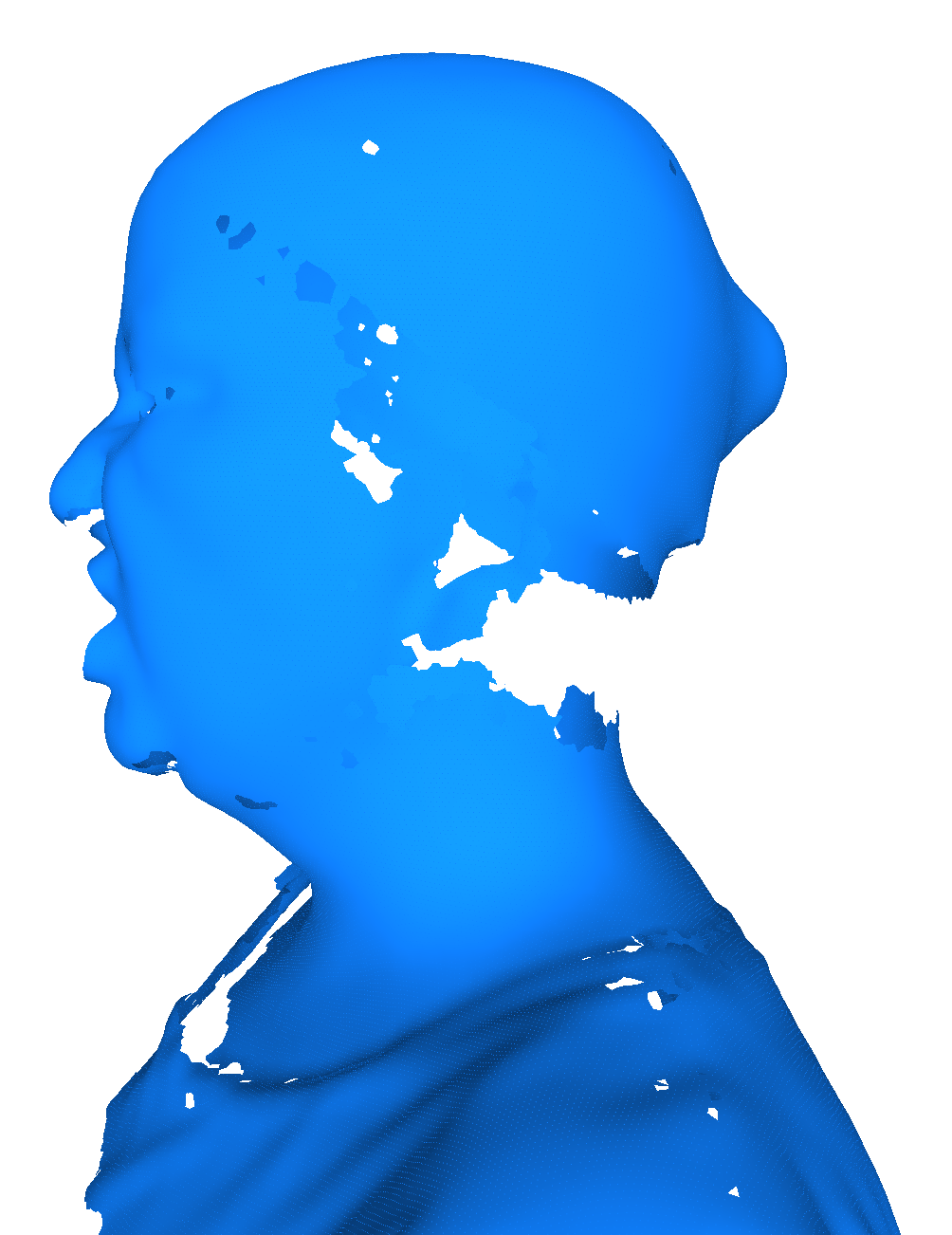}   &
                  \includegraphics[height=35mm]{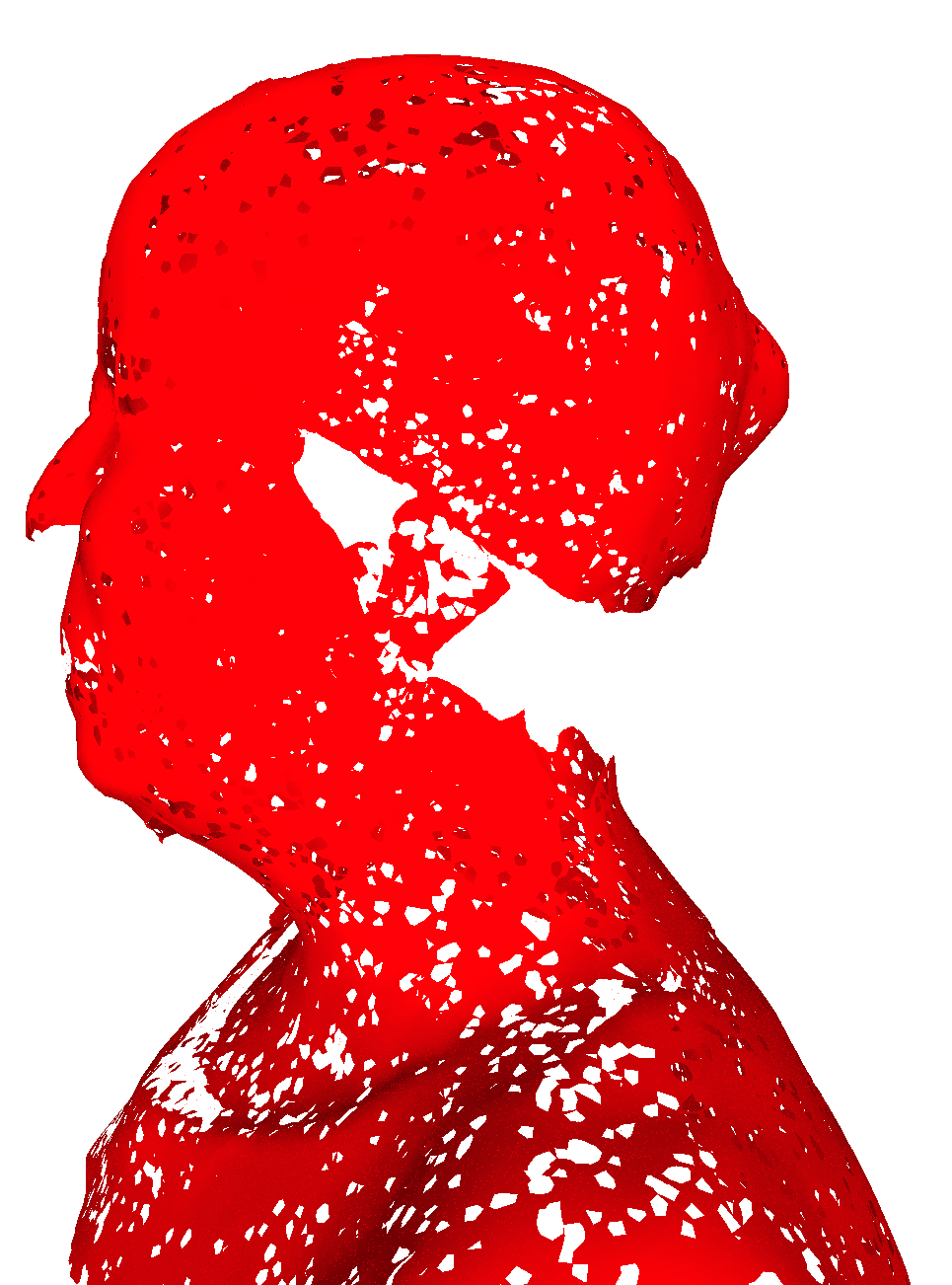}\\\bottomrule
\end{tabular}
\caption{Front and side views on NeuralQAAD reconstructions of COMA \cite{COMA}. Each head is compressed to a latent vector of size 256. Most detailed structures are reconstructed by NeuralQAAD whereas AtlasNetV2 even looses coarse structures like ears.}
\label{fig:experiments_coma}
\end{figure*}

\textbf{COMA:} COMA is the most challenging dataset as each point cloud contains almost as many points as those of D-Faust but is restricted to the facial area and, hence, captures an enormous amount of details. Again, we compress each instance into a latent code of size 256, conduct training with a batch size of 16, and make use of 128 patches for NeuralQAAD as well as 28 patches for AtlasNetV2. Surface reconstructions are shown in Figure \ref{fig:experiments_coma}.

\begin{table}[H]
\begin{tabular}{l|ccc}
                  & AtlasNetV2 & \multicolumn{2}{c}{NeuralQAAD} \\ \cline{1-4} 
Trained on        & Aug. Chamfer    & Aug. Chamfer          & QAP         \\ 
EM-kD             & 200.083         & 139.230               & 18.554   
\end{tabular}
\caption{Reconstruction EM-kD for the COMA dataset. Here, both the scalability in the number of patches as well as the QAP training procedure lead to huge performance jumps.}
\label{tab:experiments_coma}
\vspace{-7mm}
\end{table}
For both displayed instances it comes apparent that NeuralQAAD captures way more fine-grained structures than AtlasNetV2 while being less noisy. Even coarse structures like the ear in the first instance can not be recovered by AtlasNetV2. The EM-kD outcomes stated in Table \ref{tab:experiments_coma} suggest that for COMA the number of patches as well as the QAP training scheme are essential and neither alone suffices. Training NeuralQAAD on the augmented Chamfer distance achieves a 30\% better performance than AtlasNetV2. Applying the QAP training procedure further improves NeuralQAAD by a total of 90\%.
\section{Conclusion}
We introduced a new scaleable and robust point cloud autodecoder architecture called NeuralQAAD together with a novel training scheme. Its scalability origins from low level feature sharing across multiple foldable patches. Further, refraining from classical encoders makes NeuralQAAD robust to sampling. Our novel training scheme is based on two newly developed algorithms to efficiently determine an approximate solution for a specially designed quadratic assignment problem. We showed that NeuralQAAD provides better results than the previous state-of-the-art work applicable to high resolution point clouds. For this, we stated a novel scalable and fast EMD upper bound, the EM-kD. In our experiments, the EM-kD has proven to reasonably reflect visual differences between point clouds.

The next steps will be to make our approach applicable to generative models and to bridge the gap to correspondence problems. Although at first glance generative tasks seem to be a straightforward extension, preliminary results show an unstable training process. We also plan to generalize our approach to define and solve other related quadratic assignment problems.

\bibliographystyle{eg-alpha} 
\bibliography{main}
\end{document}